\pdfoutput=1

\documentclass[11pt]{article}

\usepackage[final]{acl}

\usepackage{times}
\usepackage{latexsym}

\usepackage[T1]{fontenc}

\usepackage[utf8]{inputenc}

\usepackage{microtype}

\usepackage{inconsolata}

\usepackage{graphicx}

\usepackage{xspace}
\usepackage{pifont}

\usepackage{enumitem}
\usepackage{booktabs}
\usepackage{multicol}
\usepackage{multirow}
\usepackage{caption}
\usepackage{subcaption}
\usepackage{adjustbox}
\usepackage{colortbl}
\usepackage[most]{tcolorbox}
\usepackage{makecell}
\usepackage{mdframed,lipsum}
\usepackage{enumitem}
\usepackage[capitalize]{cleveref}
\usepackage{amsfonts}
\usepackage{todonotes}

\Crefname{figure}{{Figure}}{{Figures}}
\Crefname{table}{{Table}}{{Tables}}

\setlist[itemize]{leftmargin=*}

\newtcolorbox{mybox}[1]{colback=green!5!white,colframe=green!25!black,fonttitle=\bfseries,title=#1}

\newtcolorbox{myboxcross}[1]{colback=green!5!white,colframe=green!25!black,fonttitle=\bfseries,breakable,title=#1}

\title{Sparse Neurons Carry Strong Signals of Question Ambiguity in LLMs}

\author{
  Zhuoxuan Zhang$^{1}$,
  Jinhao Duan$^{2}$,
  Edward Kim$^{2}$,
  Kaidi Xu$^{2}$ \\
  $^{1}$Brown University \quad
  $^{2}$Drexel University \\
  \texttt{zhuoxuan\_zhang@brown.edu, \{jd3734, ek826, kx46\}@drexel.edu}
}

\usepackage{amsmath,amsfonts,bm}

\def\eqref#1{equation~\ref{#1}}

\def\1{\bm{1}}

\def\vh{{\bm{h}}}

\def\vv{{\bm{v}}}
\def\vw{{\bm{w}}}
\def\vx{{\bm{x}}}

\def\vz{{\bm{z}}}

\DeclareMathAlphabet{\mathsfit}{\encodingdefault}{\sfdefault}{m}{sl}
\SetMathAlphabet{\mathsfit}{bold}{\encodingdefault}{\sfdefault}{bx}{n}

\def\vh{{\bm{h}}}
\def\vH{{\bm{H}}}

\begin{document}
\maketitle

\begin{abstract}
Ambiguity is pervasive in real-world questions, yet large language models (LLMs) often respond with confident answers rather than seeking clarification. In this work, we show that question ambiguity is linearly encoded in the internal representations of LLMs and can be both detected and controlled at the neuron level. During the model’s pre-filling stage, we identify that a small number of neurons, as few as \textbf{\textit{one}}, encode question ambiguity information. Probes trained on these \textbf{\textit{Ambiguity-Encoding Neurons}} (AENs) achieve strong performance on ambiguity detection and generalize across datasets, outperforming prompting-based and representation-based baselines. Layerwise analysis reveals that AENs emerge from shallow layers, suggesting early encoding of ambiguity signals in the model’s processing pipeline. Finally, we show that through manipulating AENs, we can control LLM's behavior from direct answering to abstention. Our findings reveal that LLMs form compact internal representations of question ambiguity, enabling interpretable and controllable behavior.
\end{abstract}

\section{Introduction}

Large Language Models (LLMs) have achieved remarkable success across various natural language processing tasks, particularly in question answering (QA). However, they often struggle with answering ambiguous questions, resulting in misleading or incorrect responses~\cite{cole2023selectively, zhang2024clamber}. Since ambiguity is common in real-world QA scenarios ~\cite{min2020ambigqa, Trienes_2019}, addressing this limitation is crucial for developing more trustworthy and reliable language systems.

Prior work has primarily addressed ambiguity from a behavioral standpoint---using prompting strategies~\cite{kuhn2022clam}, sampling-based approaches~\cite{cole2023selectively}, or training methods that encourage abstention~\cite {krasheninnikov2022assistance}. Yet, these techniques suffer from several limitations: prompt-based cues can be brittle and model-dependent; instruction tuning introduces dataset-specific biases; and decoding-time sampling is computationally costly. Crucially, these methods treat ambiguity as an input-output phenomenon, without investigating its internal representation.

\begin{figure}[t]
    \centering
    \includegraphics[width=\linewidth]{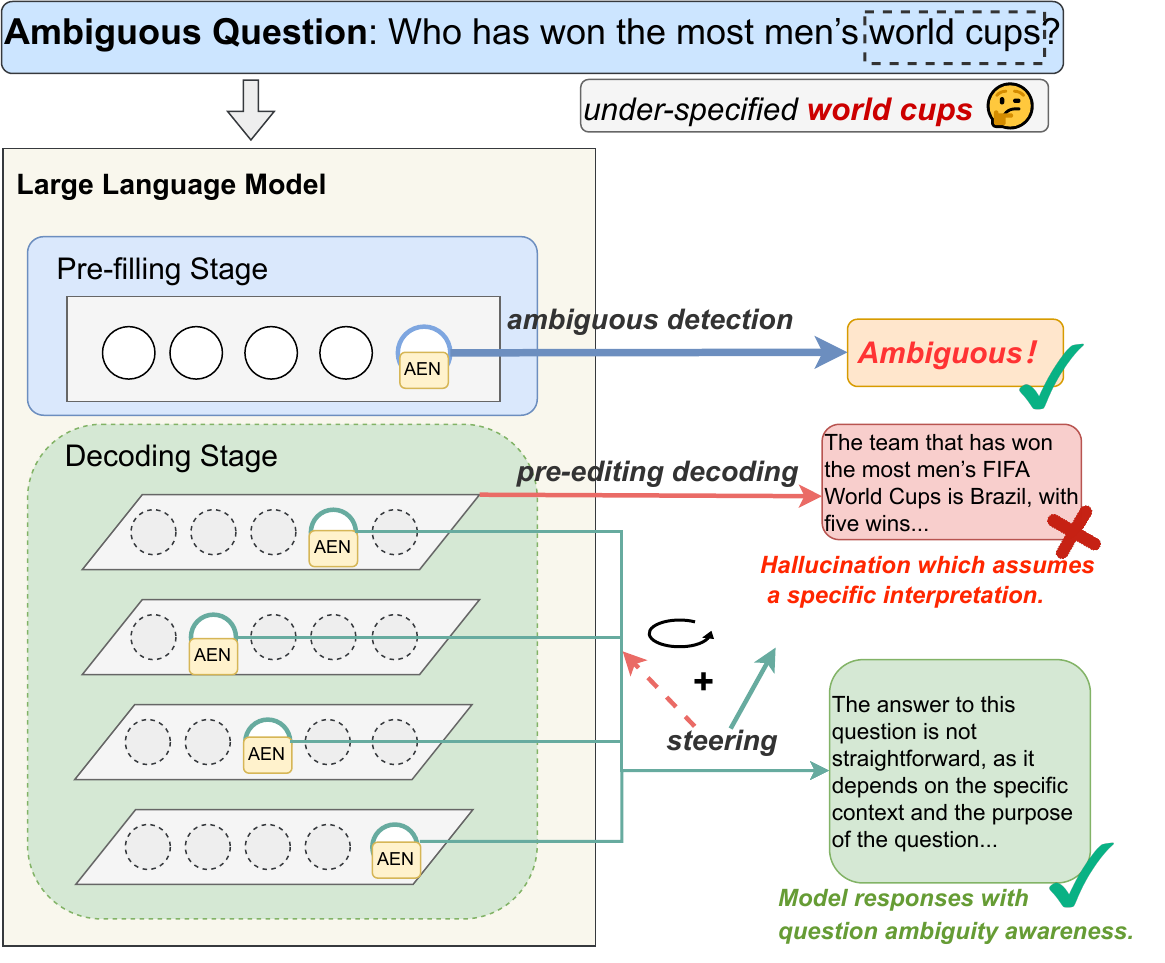}
    \caption{
    Overview of our key findings. A small set of neurons, \textit{Ambiguity-Encoding Neurons} (AENs), carry strong, linearly decodable signals of question ambiguity in LLMs. By steering the activations of these neurons alone, we causally shift model behavior from confidently answering ambiguous inputs to ambiguity-aware responses.
    }
    \label{fig:framework}
\end{figure}

In this paper, we take a fundamentally different approach: we ask how ambiguity is \emph{encoded inside the model}. Specifically, we study whether ambiguous questions are represented differently in the internal representations of LLMs from clear questions and whether these representations can be used to control LLMs' ambiguity-related behavior. We first identify signals of question ambiguity through LLM's internal activations, and then intervene on specific neurons to shift behavior from confident answering to abstention.

Our key finding is that question ambiguity is \textbf{sparsely encoded}, often in as few as \textbf{a single neuron}. We identify these \textbf{\emph{Ambiguity-Encoding Neurons}} (AENs) as predictive of question ambiguity across datasets and models, and show that steering their activations causes consistent changes in output behavior, as shown in Figure \ref{fig:framework}. These neurons emerge early in the model’s pre-filling stage, suggesting that ambiguity is recognized before generation begins.

We validate our findings in two tasks: ambiguity detection and abstention steering, across two datasets (AmbigQA ~\cite{min2020ambigqa} and SituatedQA ~\cite{zhang2021situatedqa}) and three instruction-tuned open-weight models (LLaMA 3.1 8B Instruct ~\cite{grattafiori2024llama3herdmodels}, Mistral 7B Instruct v0.3 ~\cite{jiang2023mistral7b}, and Gemma 7B IT ~\cite{team2024gemma}). Our results show that ambiguity is strongly linearly separable in internal representations, and that AENs are sufficient to detect and control this signal. These effects generalize across datasets, demonstrating the robustness of AENs.

\paragraph{Our contributions:}
\begin{itemize}
    \item We present the first neuron-level analysis of question ambiguity, showing that ambiguity is sparsely encoded in LLMs, often in as few as a single neuron, whose activation linearly separates ambiguous from unambiguous inputs.
    \item We demonstrate that steering these neurons via targeted activation manipulation causally alters model behavior, shifting responses from direct answering to abstention.
    \item We report strong empirical results across multiple instruction-tuned models and datasets, with high probe accuracy, efficient abstention control, and robust generalization.
\end{itemize}

\section{Related Work}

\noindent\textbf{Ambiguity Detection.}  
In traditional NLP, \citet{10.1007/978-3-642-14192-8_20} introduced a rule-based system for detecting ambiguities in requirements documents, while \citet{Trienes_2019} developed a classifier for unclear questions in community QA. \citet{guo2021abg} extended this by identifying ambiguity types in narratives and generating clarifying questions.  
In the LLM era, \citet{kuhn2022clam} showed that few-shot prompting enables ambiguity classification under controlled settings. \citet{krasheninnikov2022assistance} fine-tuned models to abstain or clarify when facing ambiguous queries. \citet{cole2023selectively} found that response diversity better signals ambiguity than likelihood or self-verification. \citet{zhang2024clamber} evaluated robustness across prompting strategies, revealing inconsistent model behavior. \citet{kim2024aligning} recently introduced an entropy-based metric for perceived ambiguity. We take a different approach by probing ambiguity in the internal representations of LLMs.

\noindent\textbf{Using Linear Probes to Identify Neurons.}  
Many studies have found that LLMs exhibit \emph{linear abstraction}, where latent concepts and decisions correspond to linear directions in the activation space~\citep{meng2022locating, finlayson2023causal, hernandez2022scaling, geva2022transformer}. Building on this, researchers use linear probes to identify neurons that encode specific features or behaviors. \citet{gurnee2023finding} use $k$-sparse linear probes to uncover neurons responsible for high-level features, finding increased sparsity and dedicated neurons in middle layers as model scale grows. \citet{gurnee2023language} further identifies abstract ``space'' and ``time'' neurons that generalize across contexts and entity types. SPIN~\citep{jiao2023spin} combines probing and neuron integration to improve text classification by dynamically selecting salient neurons. These works suggest that linear probes not only detect high-level structure in representations, but also serve as effective tools for neuron-level interpretability and control.

\noindent\textbf{Activation Interventions.}
Activation interventions have become a powerful tool for understanding and controlling model behavior~\citep{han2021scalecert,turner2023activation,phan2024steering, tamkin2024trustworthy}. Prior work has used this technique to steer toxicity~\citep{rimsky2024steering}, reduce hallucinations~\citep{rahn2024confident}, or control political bias~\citep{lu2024bias}. Unlike weight-based fine-tuning, activation steering provides a lightweight, reversible, and interpretable intervention. It also offers insights into the causal role of internal neurons. Several recent works further enhance the method by localizing steering to specific layers or neurons~\citep{wang2024neuronedit, stickland2024faststeer}, or decomposing the activation space~\citep{yin2024counterfact}.

\section{Method}

We investigate how question ambiguity is internally encoded and causally represented in LLMs. Our approach proceeds in two stages: (1) identifying sparse subsets of neurons that encode question ambiguity signals using linear probing, and (2) validating their functional role by assessing if targeted activation steering of these neurons causally alters the model’s behavior.

\subsection{Preliminaries}
\label{sec:prelim}
\paragraph{LLMs' Internal Representations Collection.}
Transformer-based language models process an input sequence $\vx = (\vx_1, \dots, \vx_T)$ via a series of $L$ transformer layers. At each layer $\ell \in \{1, \dots, L\}$, the model computes hidden activations $\vH^{(\ell)}(\vx) = (\vh_1^{(\ell)}, \dots, \vh_T^{(\ell)}) \in \mathbb{R}^{T \times d}$, where $\vh_t^{(\ell)}$ denotes the hidden state of token $\vx_t$ at layer $\ell$. To capture a summary of the model's internal representation during the pre-filling stage, we perform a forward pass over the prompt and aggregate the token-wise hidden states using mean pooling: $ \bar{\vh}^{(\ell)}(\vx) = \frac{1}{T} \sum_{t=1}^T \vh_t^{(\ell)} \in \mathbb{R}^d $

\paragraph{Question Ambiguity Signal.}
We define the \emph{question ambiguity signal} as an interpretable feature of a question that indicates whether it is under-specified or contextually incomplete. This signal should be detectable by humans, for example, when a question would naturally prompt a request for clarification. To model this, we use two contrastive datasets: an ambiguous set $\mathcal{D}_{\text{amb}} = \{\vx_i^{\text{amb}}\}_{i=1}^N$ composed of questions lacking key contextual information such as time or location~\citep{zhang2021situatedqa}, and a clear set $\mathcal{D}_{\text{clr}} = \{\vx_j^{\text{clr}}\}_{j=1}^N$ with sufficient context for interpretation. By comparing the model's internal representations across these sets, we aim to uncover the encoding of question ambiguity and test whether manipulating this representation can causally affect model behavior.

\paragraph{Linear Probing.}
Linear probing is a widely used technique to localize where specific information resides in a neural network by training a simple classifier to predict a labeled feature using internal activations~\citep{alain2016understanding, dalvi2019one, belrose2023eliciting, gurnee2023finding, jiao2023spin}. Given an input sequence $\vx = (\vx_1, \dots, \vx_T)$ and a transformer layer $\ell \in \{1, \dots, L\}$, the model produces hidden states $\vH^{(\ell)}(\vx) \in \mathbb{R}^{T \times d}$. These are summarized into a fixed-length representation $\vz^{(\ell)}(\vx) \in \mathbb{R}^d$ through a deterministic function (e.g., pooling or projection). A logistic regression probe then predicts a binary label via: $\hat{y}(\vx) = \sigma(\vw^\top \vz^{(\ell)}(\vx) + b), \quad \vw \in \mathbb{R}^d, \quad b \in \mathbb{R}$.
The probe is trained to minimize binary cross-entropy loss over dataset $\mathcal{D}$.
Strong probe accuracy indicates that the feature is linearly encoded in the model’s hidden states \citep{dalvi2019one}.

\paragraph{Activation Steering to alter model's behavior.}  
Activation steering is a causal intervention technique that modifies hidden activations at inference time to alter model behavior along a desired direction. Given a target vector $\vv \in \mathbb{R}^d$, which is typically derived from contrastive examples, the model’s hidden state $\vh^{(\ell)}$ at a chosen layer $\ell$ is shifted as follows: $ \tilde{\vh}^{(\ell)} = \vh^{(\ell)} + \alpha \cdot \vv$, where $\alpha$ is a scaling coefficient~\citep{turner2023activation}. To evaluate the effect of such intervention on ambiguity question handling, we partition the ambiguous question set $\mathcal{D}_{\text{amb}}$ based on the model's unmodified behavior. We label the model’s original outputs as either \emph{abstention-like} (clarifying or refusing) or \emph{direct-answering}. This yields two disjoint subsets: $\mathcal{D}^{\text{abs}}_{\text{amb}}$ for abstention-inducing examples and $\mathcal{D}^{\text{ans}}_{\text{amb}}$ for direct-answering ones.

\subsection{Linear Probing to Identify Ambiguity-Encoding Neurons}
\label{sec:linear_probe_method}

In this section, we investigate if LLM's internal representations can linearly encode question ambiguity signal. If so, how concentrated it is?

We begin by investigating whether question ambiguity is linearly encoded in the internal representations of a language model. Prior work suggests that much of a model’s understanding of an input query is formed during the pre-filling stage, and that the internal state at this point contains rich semantic information~\citep{liu2023lost, liu2023editing, mu2023towards}. In particular, the mean of token-level hidden states at a given layer has been shown to capture task-relevant signals~\citep{alain2016understanding, belrose2023eliciting, reif2019visualizing, ethayarajh2019contextual}. Motivated by these findings, we apply a logistic regression probe $\hat{y}$ to the mean activation vector $\bar{\vh}^{(\ell)}$ from the first forward pass of the model. A high classification accuracy from this probe indicates that the internal representations at layer $\ell$ encode a linearly accessible signal for question ambiguity. We denote this probe's performance as $\text{Acc}_{\text{full}}$.

We then identify the neurons most responsible for encoding question ambiguity by analyzing the dimensions of $\hat{y}$ that contribute most to the probe’s prediction. Specifically, we examine the learned weight vector $\vw$ of the trained probe to locate the most influential dimensions. We rank each dimension $i$ by the absolute value of its weight $|\vw_i|$, which serves as a proxy for salience ~\citep{tibshirani1996regression, guyon2003introduction, ng2004feature}. The index set of the top-$k$ highest-ranked neurons is denoted as $\mathcal{S}_k$. Following prior work that employs noise injection to study network's functionality ~\citep{levi2022noise, mahadevan2021machine, beinecke2021comparative}, we assess the functional role of top-$k$ neurons by injecting Gaussian noise into their corresponding dimensions. For each $i \in \mathcal{S}_k$, we perturb the $i$-th coordinate of the hidden representation as follows:
\[
\tilde{\vh}_i^{(\ell)} =
\begin{cases}
\bar{\vh}_i^{(\ell)} + \epsilon_i, & \text{if } i \in \mathcal{S}_k \\
\bar{\vh}_i^{(\ell)}, & \text{otherwise}
\end{cases}, \quad \epsilon_i \sim \mathcal{N}(0, \sigma^2)
\]
We then compute the classification accuracy of the linear probe on the perturbed representation and define the resulting accuracy degradation as: $ \Delta_{\text{acc}}(k) := \text{Acc}_{\text{full}} - \text{Acc}_{\text{perturbed}(\mathcal{S}_k)}$.
We designate $\mathcal{S}_k$ as the set of \textbf{\emph{Ambiguity-Encoding Neurons (AENs)}} when this drop is maximized across varying values of $k$, indicating that these dimensions are critical for encoding ambiguity signals.

To further validate that the AENs capture sufficient predictive signal, we adopt a sparse probing approach following~\citet{dalvi2019one}, training a logistic regression classifier restricted only to the top-$k$ dimensions in $\mathcal{S}_k$:
\[
\hat{y}_{\text{AENs}} = \sigma\left(\vw_{\mathcal{S}_k}^\top \bar{\vh}_{\mathcal{S}_k}^{(\ell)} + b\right), \quad \bar{\vh}_{\mathcal{S}_k}^{(\ell)} \in \mathbb{R}^k
\]
Despite their extreme sparsity, these AENs probes achieve accuracy close to $\text{Acc}_{\text{full}}$, providing strong evidence that the selected neurons alone carry sufficient information to predict ambiguity. This confirms that the ambiguity signal is not diffusely distributed, but instead concentrated in a compact, interpretable subspace.

\subsection{Causal Neuron-Level Steering}
\label{sec:causal_ctivation_steering}

To assess whether the identified AENs encode functionally meaningful representations of ambiguity, we test their causal influence on model behavior through activation steering~\citep{wang2024neuronedit, stickland2024faststeer, yin2024counterfact}. Specifically, we investigate whether modifying such a small subset of neurons can reliably shift model outputs from direct answers to abstentions.

To construct steering directions, we adopt the contrastive representation method introduced by~\citet{lee2024programming}, which involves mean-centering and applying principal component analysis (PCA) over sets of hidden activations corresponding to different behaviors. Specifically, we define $\mathcal{D}^{+} := \mathcal{D}_{\text{amb}}^{\text{abs}}$ as ambiguous prompts that originally elicited \emph{abstention} behavior (e.g., clarification or refusal), and $\mathcal{D}^{-} := \mathcal{D}_{\text{clr}}$ as prompts that received direct answers. To ensure consistent decoding across examples, we follow~\citet{lee2024programming} by appending a suffix to each input to reinforce the target response style. 

For each example $\vx$, we compute the hidden representation $\bar{\vh}^{(\ell)}$ by mean-pooling over all token activations in the input sequence at layer $\ell$. Then we define $\vH_+^{(\ell)} = \{ \bar{\vh}^{(\ell)}(\vx) \mid \vx \in \mathcal{D}^+ \}$ and $\vH_-^{(\ell)} = \{ \bar{\vh}^{(\ell)}(\vx) \mid \vx \in \mathcal{D}^- \}$ as the hidden states for abstention and answering examples, respectively. To compute the steering direction, we first calculate the mean of both groups:
\[
\boldsymbol{\mu}^{(\ell)} = \frac{1}{2} \left( \frac{1}{|\mathcal{D}^{+}|} \sum_{\vx \in \mathcal{D}^{+}} \bar{\vh}^{(\ell)}(\vx) + \frac{1}{|\mathcal{D}^{-}|} \sum_{\vx \in \mathcal{D}^{-}} \bar{\vh}^{(\ell)}(\vx) \right)
\]
We then mean-center both sets and concatenate them as input to PCA:
\[
\boldsymbol{\Delta}^{(\ell)} = \text{PCA}_1\left(\left[\vH_+^{(\ell)} - \boldsymbol{\mu}^{(\ell)}; \, \vH_-^{(\ell)} - \boldsymbol{\mu}^{(\ell)} \right]\right)
\]
The first principal component $\boldsymbol{\Delta}^{(\ell)}$ captures the dominant contrastive direction between abstention and answering behaviors for each layer $\ell$.

At test time, for an ambiguous prompt $\vx \in \mathcal{D}_{\text{amb}}^{\text{ans}}$, we apply steering as:
\[
\tilde{\vh}^{(\ell)}(\vx) = \bar{\vh}^{(\ell)}(\vx) + \alpha \cdot \left(\text{Mask}^{(\ell)} \odot \boldsymbol{\Delta}^{(\ell)}\right)
\]
where $\alpha$ is a scaling factor, $\text{Mask}^{(\ell)} \in \{0, 1\}^d$ specifies the modified neurons, and $\odot$ is elementwise multiplication.

We experiment with three neuron selection strategies for steering: \textbf{full vector steering}, which modifies all neurons ($\text{Mask}^{(\ell)} = \mathbf{1}$); \textbf{AENs steering}, which modifies only the $k$ neurons in $\mathcal{S}_k$ identified as Ambiguity-Encoding Neurons in Section~\ref{sec:linear_probe_method}; and \textbf{top-$p$ neuron steering}, which modifies the top $p \in \{50, 100\}$ neurons ranked by the magnitude of probe weights $|\vw_i|$.

Steering is applied to ambiguous prompts in $\mathcal{D}_{\text{amb}}^{\text{ans}}$, which initially elicited direct answers. We assess the intervention’s effectiveness by measuring whether the model's responses shift toward abstention.

\section{Experiments}

Our experiments address four core questions: (1) whether ambiguity is linearly decodable, by testing if a probe trained on hidden states can reliably distinguish ambiguous from unambiguous questions; (2) whether a small set of neurons contains strong question ambiguity signal; (3) whether these neurons are sufficient for generalizable detection, by comparing AENs probes to full-vector probes and existing ambiguity detection baselines across datasets; and (4) whether AENs causally control model behavior, by evaluating if activation steering on these neurons shifts model outputs from confidently answering to abstention.

\subsection{Setup}
\label{sec:setup}
\paragraph{Models.}
We evaluate three open-weight instruction-tuned language models: \textbf{LLaMA 3.1 8B Instruct}~\citep{grattafiori2024llama3herdmodels}, \textbf{Mistral 7B Instruct v0.3}~\citep{jiang2023mistral7b}, and \textbf{Gemma 7B IT}~\citep{team2024gemma}. For brevity, we often refer to these models as \textbf{LLaMA 3.1 8B}, \textbf{Mistral 7B}, and \textbf{Gemma 7B} in the rest of the paper. All generations use temperature 0.1 for consistency.

\paragraph{Datasets.}
We use \textbf{AmbigQA}~\citep{min2020ambigqa} and \textbf{SituatedQA}~\citep{zhang2021situatedqa} to build contrastive splits. We construct paired examples for ambiguity detection: \(\mathcal{D}_{\text{probe}} = \{(\vx_i^{\text{amb}}, \vx_i^{\text{clr}})\}_{i=1}^N\).
Each set is randomly shuffled and split into 400 training and 1000 testing examples per class. These are used to train and evaluate linear probes.

Separately, for activation steering, we partition ambiguous prompts based on model behavior. A pretrained LLM-as-judge labels responses as either \textit{abstention} (clarification or refusal) or \textit{direct answer}, yielding: \(\mathcal{D}_{\text{amb}}^{\text{abs}}, \quad \mathcal{D}_{\text{amb}}^{\text{ans}}\). We construct steering vectors using 100 abstention examples from $\mathcal{D}_{\text{amb}}^{\text{abs}}$ and 100 clear examples from $\mathcal{D}_{\text{clr}}$, and evaluate the resulting behavior shift on 500 ambiguous prompts from $\mathcal{D}_{\text{amb}}^{\text{ans}}$. Details of datasets and LLM-as-judge implementation are provided in Appendix~\ref{appendix:dataset-details}.

\paragraph{Feature Extraction.}
For each input, we extract hidden states from layer $\ell$ and mean-pool over the sequence as stated in Section \ref{sec:prelim}. Unless otherwise stated, we use $\ell = 14$ as the default probing layer. Layerwise results appear in Section~\ref{sec:layerwise_analysis}.

\paragraph{Ambiguity Detection Baselines.}
We compare against prompting and representation-based methods:
CLAM~\citep{kuhn2022clam}, CLAMBER~\citep{zhang2024clamber}, and INFOGAIN~\citep{kim2024aligning}. Prompt templates and implementation details appear in Appendix~\ref{sec:baseline_implementation}.

\begin{figure*}[t]
    \centering
    \includegraphics[width=\textwidth]{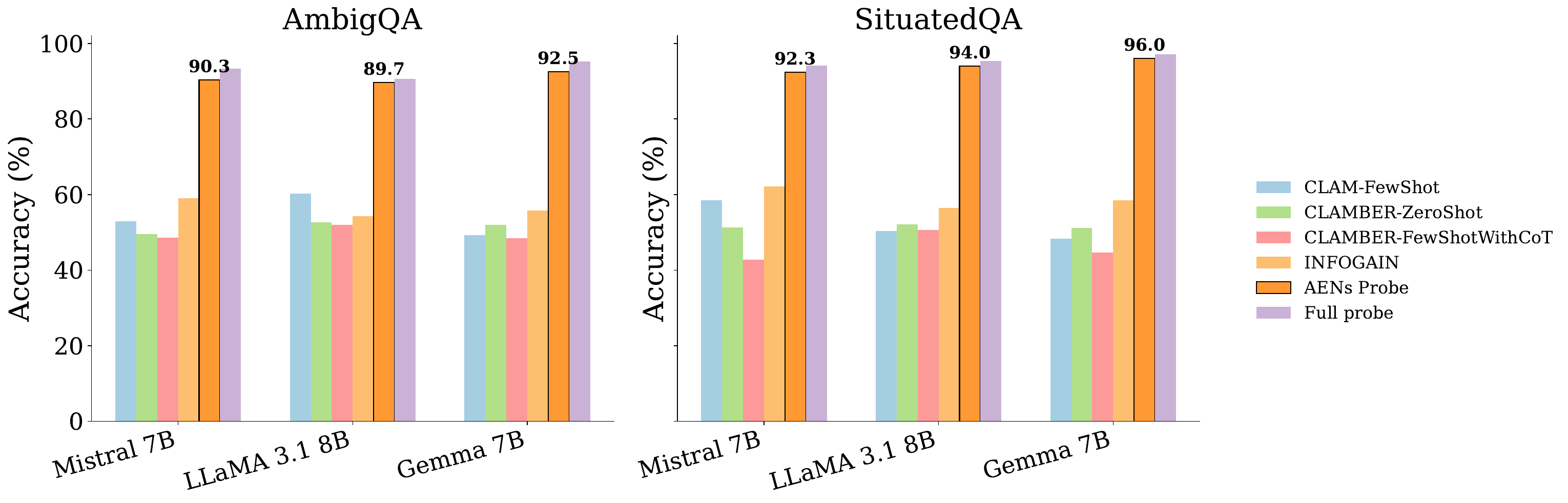}
    \caption{Accuracy of AENs probes across AmbigQA and SituatedQA. AENs probes perform comparably to full probe models and outperform baselines.}
    \label{fig:probe-accuracy}
\end{figure*}

\subsection{Linear Probings to Locate and Validate Ambiguity-Encoding Neurons}
\begin{table}[t]
\centering
\adjustbox{width=\linewidth}{
\begin{tabular}{lcccc}
\toprule
& \textbf{Accuracy} & \textbf{Precision} & \textbf{Recall} & \textbf{F1} \\
\toprule
\multicolumn{5}{c}{\textsc{AmbigQA}} \\
\midrule
Mistral 7B & 93.30 & 93.48 & 93.30 & 93.29 \\
LLaMA 3.1 8B & 90.65 & 91.79 & 90.65 & 90.59 \\
Gemma 7B & 95.25 & 95.53 & 95.25 & 95.24 \\
\midrule
\multicolumn{5}{c}{\textsc{SituatedQA}} \\
\midrule
Mistral 7B & 94.14 & 94.57 & 94.15 & 94.14 \\
LLaMA 3.1 8B & 95.40 & 95.74 & 95.40 & 95.39 \\
Gemma 7B & 97.10 & 97.12 & 97.10 & 97.10 \\
\bottomrule
\end{tabular}
}
\caption{Macro-averaged accuracy, precision, recall, and F1 of linear probes trained on AmbigQA and SituatedQA.}
\label{tab:probe-evaluation}
\end{table}

\begin{table}[t]
\centering
\small
\adjustbox{width=\linewidth}{
\begin{tabular}{llp{0.6\linewidth}}
\toprule
\textbf{Model} & \textbf{Dataset} & \textbf{Top-5 Neurons (by $|w|$)} \\
\midrule
\multirow{2}{*}{Mistral 7B}
  & \textsc{AmbigQA}     & \textbf{2070}, 3240, 2043, 1909, 1372 \\
  & \textsc{SituatedQA}  & \textbf{2070}, 2388, 2078, 53, 2083 \\
\midrule
\multirow{2}{*}{LLaMA 3.1 8B}
  & \textsc{AmbigQA}     & \textbf{788}, \textbf{1384}, \textbf{4062}, 4055, 1298 \\
  & \textsc{SituatedQA}  & \textbf{788}, \textbf{1384}, \textbf{4062}, 4055, 3231 \\
\midrule
\multirow{2}{*}{Gemma 7B}
  & \textsc{AmbigQA}     & \textbf{1995}, 1963, 1496, 1288, 2217 \\
  & \textsc{SituatedQA}  & \textbf{1995}, 1258, 1355, 1884, 155 \\
\bottomrule
\end{tabular}
}
\caption{
Top-5 most important neurons by probe weight for each model on AmbigQA and SituatedQA.
\textbf{Bolded neurons} indicate AENs shared across both datasets for the same model.
}
\label{tab:top-neurons-all}
\end{table}


\label{sec:identifying-Ambiguity-Encoding Neurons}

We first ask whether ambiguity is linearly accessible in the model’s internal representations. As shown in Table~\ref{tab:probe-evaluation}, probes achieve high accuracy across both datasets and all models, demonstrating strong linear separability.

Then we investigate \textit{where this signal is encoded, and how concentrated it is?}

\vspace{0.5em}
\noindent\textbf{Locating Ambiguity-Encoding Neurons.}  
To identify where ambiguity is encoded in the model, we rank hidden dimensions by the magnitude of their corresponding weights \( |w_i| \) from a trained linear probe. This highlights the most influential dimensions for classification. To validate their importance, we iteratively inject Gaussian noise into the top-$k$ dimensions and measure the resulting drop in classification accuracy. A sharp accuracy decline indicates that these dimensions are critical for encoding the ambiguity signal.

\begin{figure}[t]
    \centering
    \includegraphics[width=\linewidth]{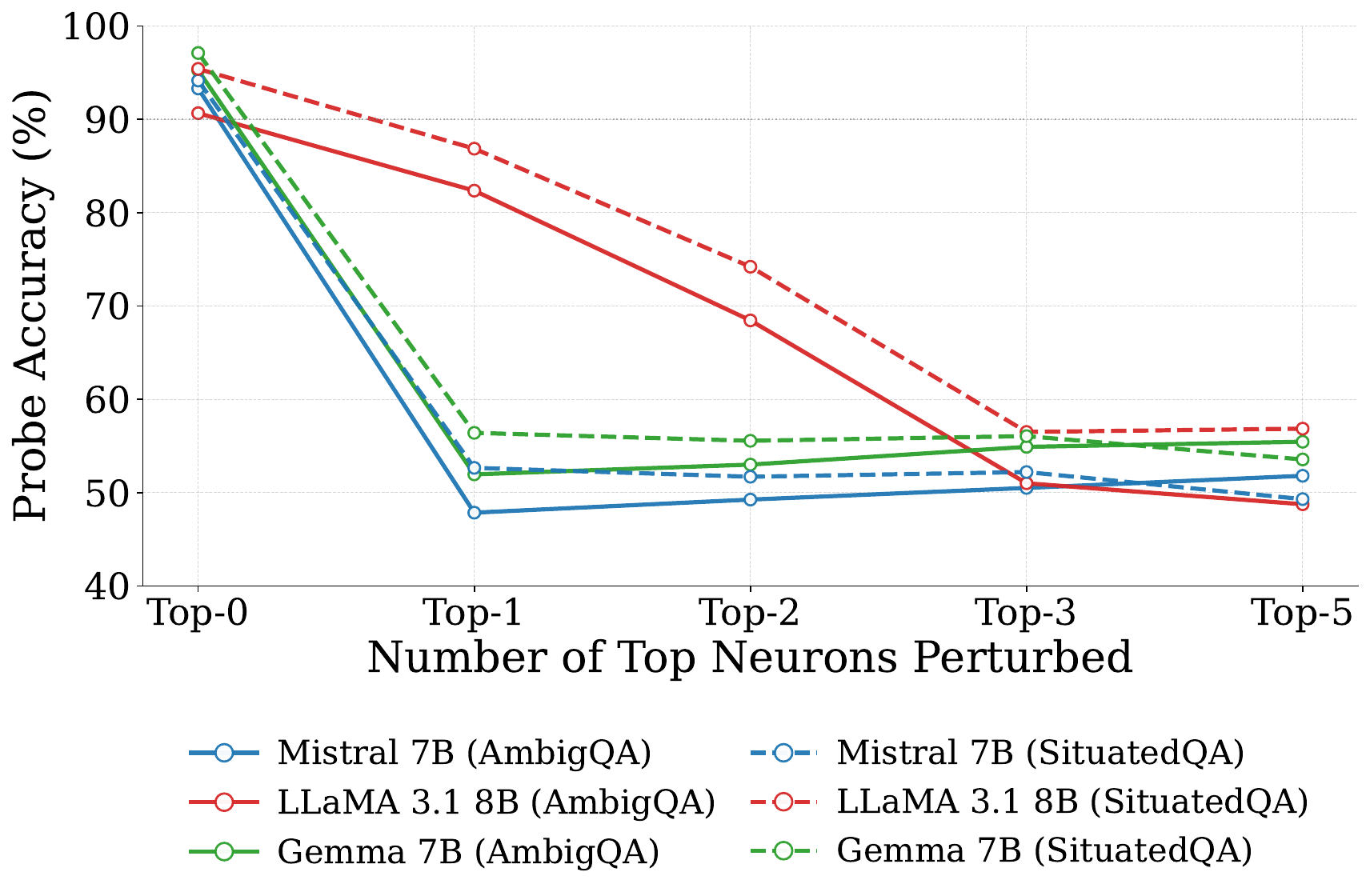}
    \caption{Probe accuracy after perturbing top-$k$ most predictive neurons. Even a small number of altered dimensions causes sharp performance drops, showing sparsity of the ambiguity signal.}
    \label{fig:accuracy-drop}
\end{figure}

Figure~\ref{fig:accuracy-drop} shows that perturbing even a few neurons can sharply reduce classification accuracy. We identify 1 such neuron for Mistral 7B and Gemma 7B, and 3 for LLaMA~3.1 8B. We refer to these highly influential neurons as \textit{\textbf{Ambiguity-Encoding Neurons (AENs)}}, as they contain predictive signals for linearly separating ambiguous from unambiguous inputs in the probe classifier. This extreme sparsity suggests that ambiguity is not diffusely encoded, but instead concentrated in a small, identifiable subspace.

Notably, the same neuron indices are identified as AENs across both AmbigQA and SituatedQA for each model (Table~\ref{tab:top-neurons-all}, bolded), suggesting that ambiguity is encoded in a consistent, model-specific subspace that generalizes across domains.

\vspace{0.5em}
\noindent\textbf{Validating Ambiguity-Encoding Neurons.}
To validate that the identified AENs genuinely encode question ambiguity, we retrain logistic regression classifiers using only AENs. We refer to these classifiers as \textit{AENs probes}. Despite their extreme sparsity, AENs probes achieve strong predictive performance. As shown in Figure~\ref{fig:probe-accuracy}, they match or exceed the accuracy of prior ambiguity detection baselines and approach the performance of full-dimension probes \textit{full probes}, which use the entire hidden representation. This provides compelling evidence that AENs concentrate the core signal needed to distinguish ambiguous from unambiguous questions. Full numerical results, including F1 scores and comparisons with all baselines, are provided in Appendix~\ref{appendix:probe-table}.

\begin{figure}[t]
    \centering
    \includegraphics[width=0.9\linewidth]{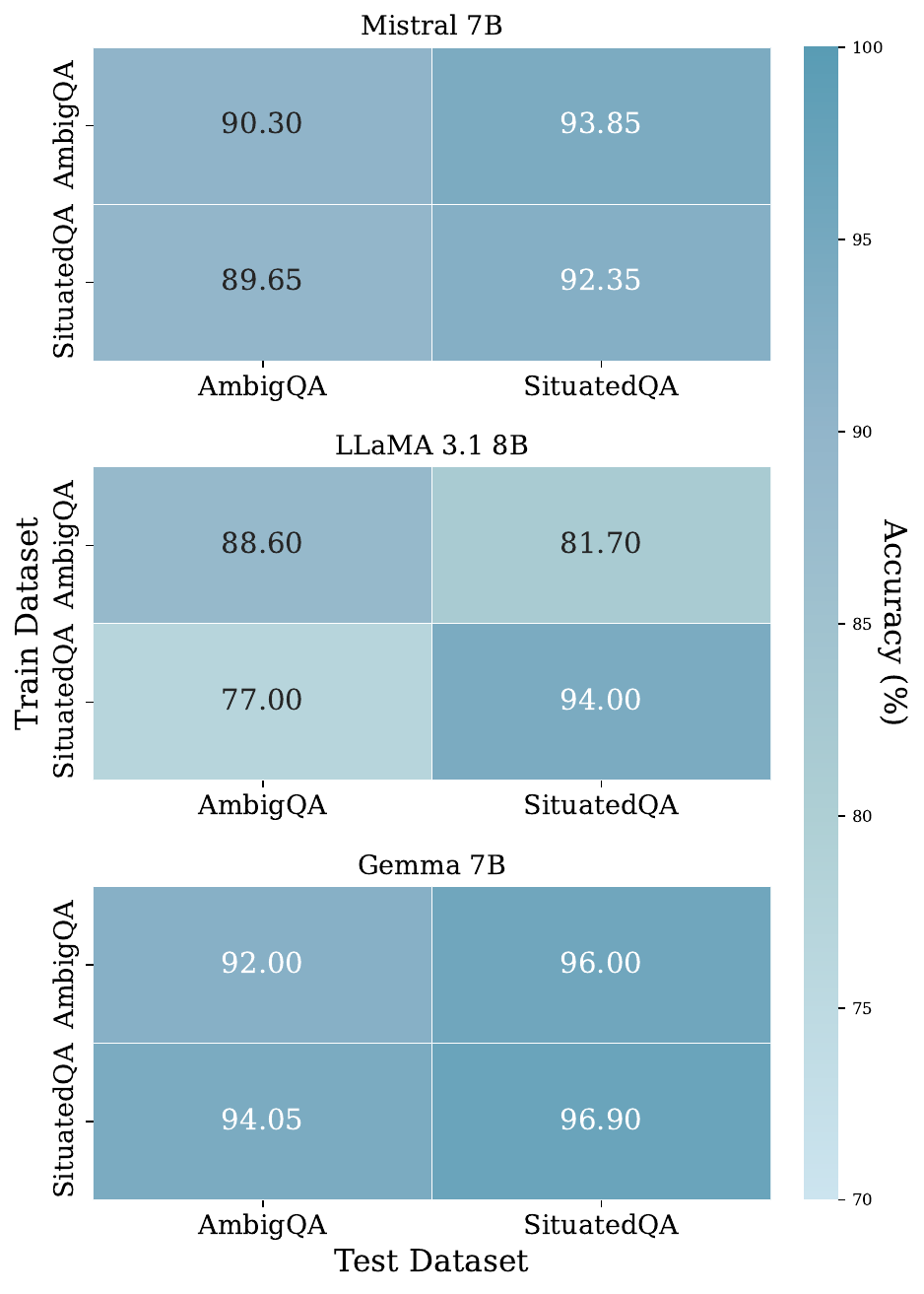}
    \caption{
    Cross-domain confusion matrices for AENs probes on each model. Values reflect classification accuracy (\%). Probes generalize robustly across datasets.
    }
    \label{fig:cross_domain_confusion_matrix}
\end{figure}

We further assess the robustness of these representations through cross-domain generalization. Specifically, we train AENs probes on one dataset (e.g., AmbigQA) and evaluate on another (e.g., SituatedQA). As shown in Figure~\ref{fig:cross_domain_confusion_matrix}, AENs probes generalize well across domains, supporting the view that these neurons encode domain-invariant features of ambiguity.

\subsection{Causal Neuron-Level Steering}
\label{sec:activation_steering}
\begin{table}[t]
\centering
\small
\adjustbox{width=\linewidth}{
\begin{tabular}{llccc}
\toprule
\textbf{Dataset} & \textbf{Steering Type} & \textbf{Mistral 7B} & \textbf{LLaMA 3.1 8B} & \textbf{Gemma 7B} \\
\midrule
\multirow{4}{*}{\textsc{AmbigQA}} 
& AENs              & 18.0 & 52.0 & 13.2 \\
& Top 50 Neurons    & 27.4 & 54.6 & 20.0 \\
& Top 100 Neurons   & 38.4 & 58.2 & 28.8 \\
& Full Vector       & 68.8 & 62.8 & 53.6 \\
\midrule
\multirow{4}{*}{\textsc{SituatedQA}} 
& AENs              & 23.8 & 50.4 & 11.6 \\
& Top 50 Neurons    & 32.8 & 62.6 & 16.0 \\
& Top 100 Neurons   & 35.4 & 74.0 & 17.6 \\
& Full Vector       & 73.6 & 93.2 & 56.8 \\
\bottomrule
\end{tabular}
}
\vspace{0.5em}
\caption{Abstention rate (\%) under different steering configurations. Experiments are conducted over a test set where LLMs always directly answer the question, i.e., the vanilla abstention rate is 0\%.}
\label{tab:steering-super-vs-topk}
\end{table}

\begin{figure*}[t]
    \centering
    \includegraphics[width=1\textwidth]{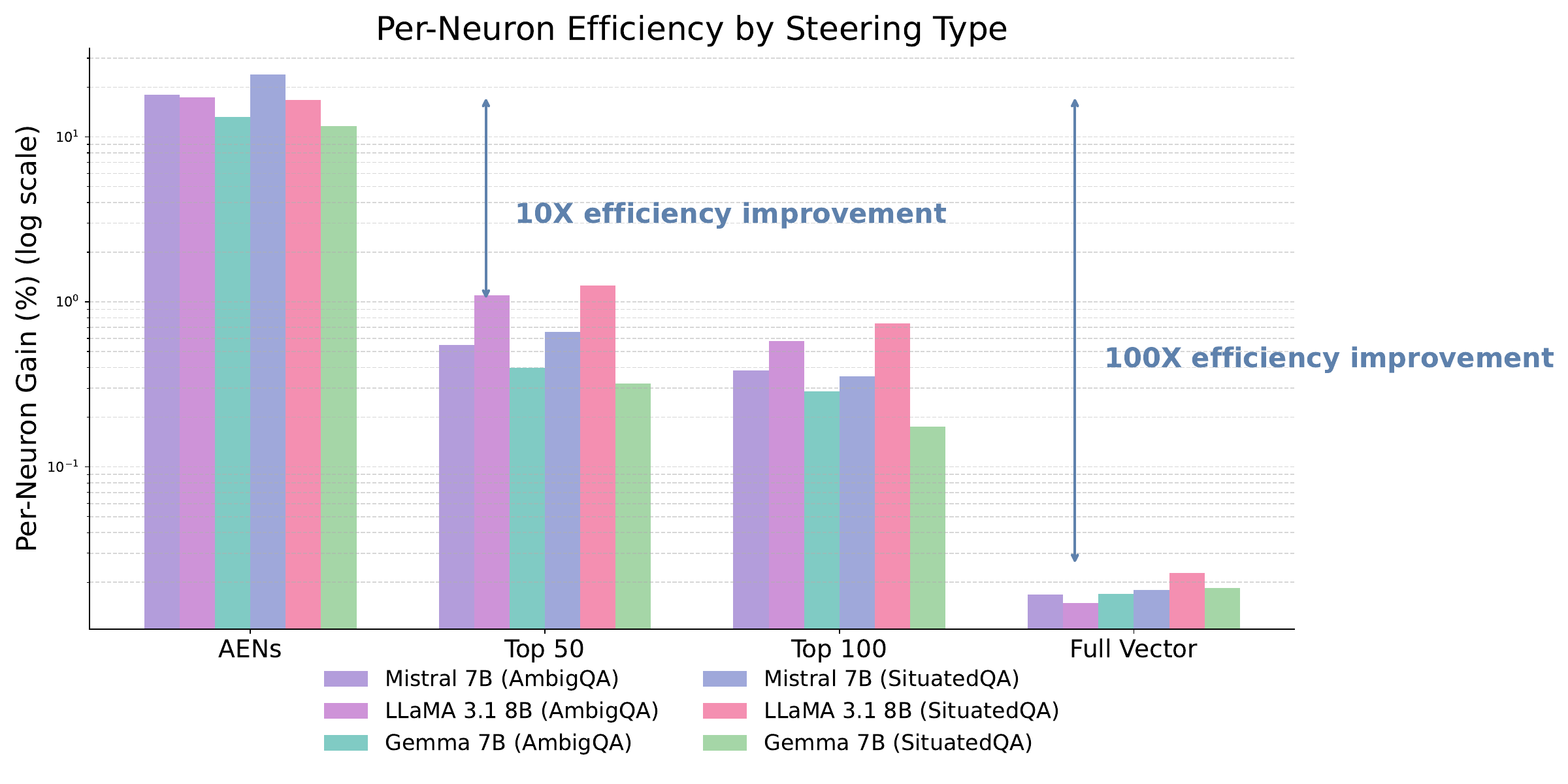}
    \caption{Per-neuron gain (\% increase in abstention per neuron) under each steering method. AENs steerings consistently show the highest efficiency across models and datasets.}
    \label{fig:per-neuron-efficiency}
\end{figure*}

\begin{figure}[t]
    \centering
    \includegraphics[width=\linewidth]{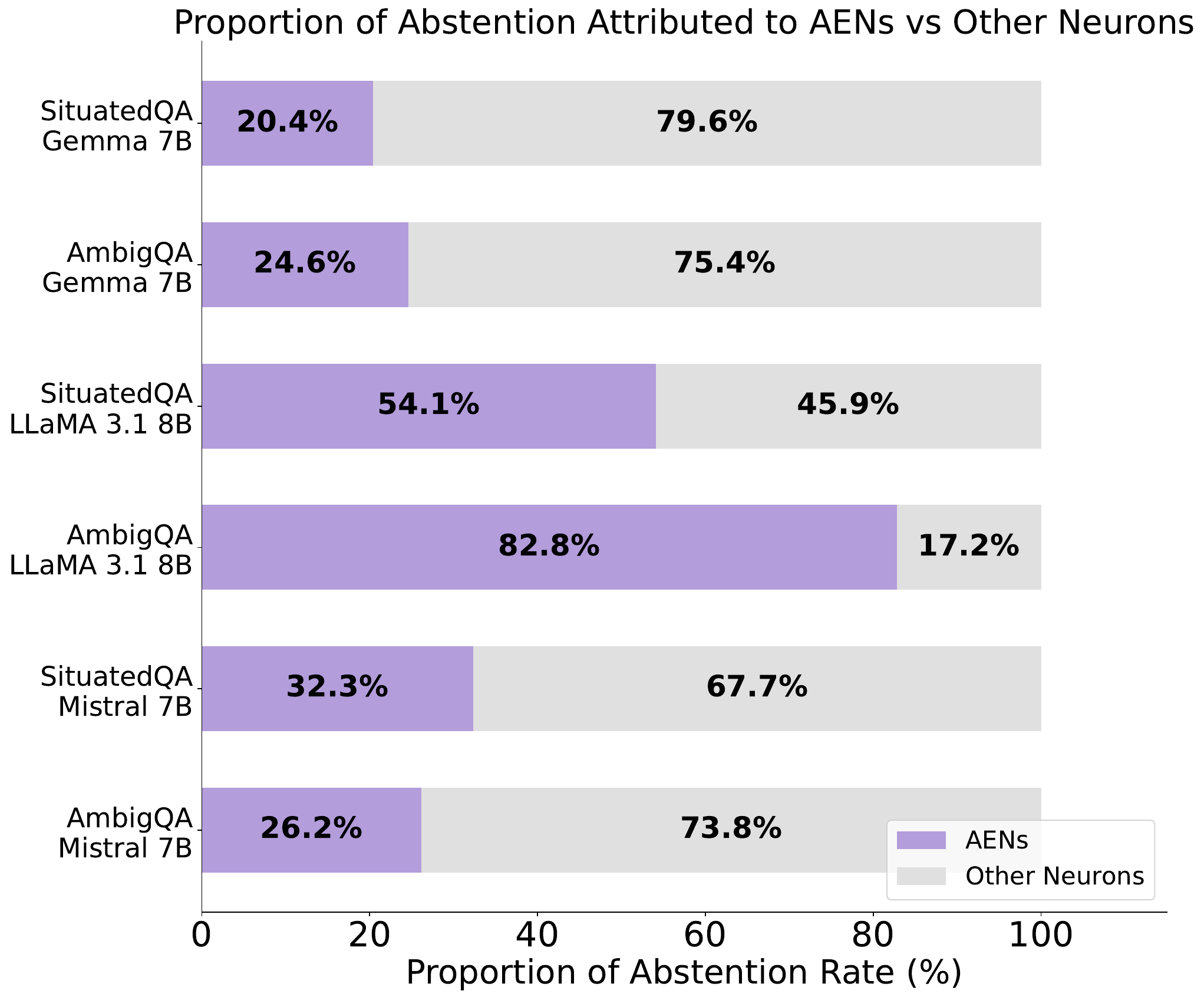}
    \caption{Bar charts showing the proportion of abstention rate achieved by AENs steering to full vector steering. AENs capture a great portion of the steering effect.}
    \label{fig:donut-steering-contribution}
\end{figure}

To validate whether AENs encode meaningful representations of the ambiguity signal, we apply \textbf{activation steering} to modify the hidden states in a targeted manner, aiming to shift the model’s behavior from answering ambiguous questions to abstention.

We follow Section \ref{sec:causal_ctivation_steering} to construct a behavior direction $\Delta^{(\ell)}$ and apply to ambiguous inputs from $\mathcal{D}_{\text{amb}}^{\text{ans}}$, and we use an LLM-as-judge (Appendix~\ref{appendix:llm-judge}) to evaluate whether outputs exhibit abstention.

We compare three steering strategies: (1) \textbf{AENs steering}, which targets the small set of neurons identified in Section~\ref{sec:identifying-Ambiguity-Encoding Neurons}; (2) \textbf{Top-$k$ neurons steering}, which modifies the top 50 or 100 neurons ranked by probe weight magnitude $|w_i|$; and (3) \textbf{Full vector steering}, which applies the intervention across all hidden dimensions.

\paragraph{AENs Are Causally Effective and Efficient.}
As shown in Table~\ref{tab:steering-super-vs-topk}, steering only a few AENs leads to a substantial shift in behavior. For instance, \textsc{LLaMA 3.1 8B} Instruct reaches 52.0\% abstention on \textsc{AmbigQA} with just 3 neurons (AENs), nearly matching the 58.2\% from steering 100 neurons. 

We quantify this in terms of \textbf{per-neuron gain}, computed as the additional abstention rate per added neuron. As visualized in Figure~\ref{fig:per-neuron-efficiency}, AENs steering consistently outperform all other methods by more than $10\times$ to $100\times$ in efficiency across all models and datasets.

\paragraph{Top-$k$ and Full-Vector Steering Show Diminishing Returns.}
While top-100 and full-vector steering produce higher absolute abstention rates, they do so at far greater cost. For example, steering all neurons in \textsc{Gemma 7B IT} yields 56.8\% abstention on \textsc{SituatedQA}, but steering just AEN (one neuron) achieves 11.6\%. AENs capture great behavioral effects.

\paragraph{AENs Capture the Majority of the Full Steering Effect.}
Figure~\ref{fig:donut-steering-contribution} shows the proportion of the full-vector abstention effect explained by AENs. In \textsc{LLaMA 3.1 8B Instruct}, AENs have over 50\% of the full effect on both datasets, despite modifying just 3 out of thousands of neurons. This highlights their disproportionately large causal influence.

\paragraph{Qualitative Analysis of AENs Steering}
\label{sec:qualitative_steering}
To illustrate the behavioral effect of AENs steering, we present model response examples before and after intervention, as shown in Appendix Table~\ref{tab:qualitative-aen}. We demonstrate that models can give reasonable abstention answers to questions.

\subsection{Ablation Studies}
\subsubsection{Layerwise Analysis: Emergence of Question Ambiguity Signal}
\label{sec:layerwise_analysis}

We perform a layerwise probing analysis across all transformer layers to investigate where question ambiguity signals emerge within the model. We train two logistic regression classifiers at each layer: one using the full neurons and another using only the AENs. As shown in Appendix~\ref{appendix:layerwise}, probe accuracy rises rapidly in early layers and saturates before Layer~5 across all three models. For example, in \textsc{Gemma 7B IT}, AENs probe accuracy surpasses 90\% as early as Layer~2. This suggests that ambiguity becomes linearly accessible within the shallow layers of the model and is sparsely encoded.

\subsubsection{Distributional Shift: AENs vs.\ Other Neurons}
\label{sec:distributional_shift}

To further investigate why AENs are especially effective for ambiguity detection, we analyze the statistical behavior of their activations under ambiguous and clear prompts. Specifically, we compare the distribution of activations for AENs to those of non-AEN neurons. We find that AENs exhibit much larger differences in activation means between ambiguous and clear inputs than other neurons. Details can be found in Appendix \ref{appendix:distribution_shift}.

\subsubsection{Cross-Domain Evaluation of AENs Steering}
\label{sec:cross-domain-steering}

To test the robustness of AENs steering, we evaluate whether ambiguity steering vectors constructed from one dataset transfer to another. Specifically, we extract the steering vector $\vv$ using AmbigQA, then apply it using AENs neurons only in SituatedQA, and vice versa. We find that AENs steering retains strong effectiveness across domains, indicating that the ambiguity signal encoded by these neurons is not dataset-specific, and thus shows that AENs capture a semantically grounded and transferable representation of question ambiguity. Full results are reported in Appendix~\ref{appendix:cross_domain}.

\subsubsection{Unintended Side Effects of AENs Steering}
\label{sec:side-effects}

We assess whether AENs steering introduces any undesirable behaviors. Since AENs steering is applied only when a question is classified as ambiguous, we evaluate its potential side effects in two scenarios: (1) false positives on clear questions, and (2) disruption of existing abstention behavior on ambiguous questions.

To evaluate false positives, we apply our trained AENs classifier to 1,000 questions from TriviaQA \cite{joshi-etal-2017-triviaqa}, a factual QA dataset with mostly unambiguous questions. All models maintain high classification accuracy, as shown in Table \ref{tab:side-effects-triviaqa}, suggesting that AENs are unlikely to misfire on clear inputs.

We then evaluate ambiguous cases where the base models abstained and test whether AENs steering meaningfully changes this behavior. We use an LLM-as-judge to assess whether abstention behavior is preserved. We define \textit{abstention consistency} as the proportion of instances where abstention remains unchanged after steering. As shown in Table~\ref{tab:side-effects-abstention}, consistency stays above 92\% across all models and datasets, indicating that AENs steering preserves the model’s original abstention and meaning.

\begin{table}[t]
\centering
\renewcommand{\arraystretch}{1.15}
\setlength{\tabcolsep}{12pt}
\adjustbox{width=\linewidth}{
\begin{tabular}{lccc}
\toprule
\textbf{Dataset} & \textbf{LLaMA~3.1~8B} & \textbf{Mistral~7B} & \textbf{Gemma~7B} \\
\midrule
AmbigQA     & 89.9\% & 98.5\% & 89.2\% \\
SituatedQA  & 90.6\% & 96.0\% & 88.7\% \\
\bottomrule
\end{tabular}}
\caption{
AEN-based classifier accuracy on 1,000 TriviaQA examples. Classifier trained on AmbigQA or SituatedQA using AENs. High accuracy indicates low false positive rate.
}
\label{tab:side-effects-triviaqa}
\end{table}

\vspace{0.75em}

\begin{table}[t]
\centering
\renewcommand{\arraystretch}{1.15}
\setlength{\tabcolsep}{12pt}
\adjustbox{width=\linewidth}{
\begin{tabular}{lccc}
\toprule
\textbf{Dataset} & \textbf{LLaMA~3.1~8B} & \textbf{Mistral~7B} & \textbf{Gemma~7B} \\
\midrule
AmbigQA     & 98.8\% & 94.6\% & 97.0\% \\
SituatedQA  & 95.2\% & 92.6\% & 95.8\% \\
\bottomrule
\end{tabular}}
\caption{
Abstention consistency post-AEN steering: Percentage of ambiguous examples where the model's original abstention behavior is preserved. High values indicate that AEN steering is minimally disruptive.
}
\label{tab:side-effects-abstention}
\end{table}

\subsubsection{Reverse Steering: From Abstention to Direct Answering}
\label{sec:reverse-steering}

We investigate whether AENs support bidirectional control by steering in the reverse direction, i.e., converting abstentions into direct answers. We construct a set of 500 ambiguous questions per dataset where the models abstained and apply the inverted steering direction ($-\vv$) using the same AENs identified earlier. We then evaluate the result following the LLM-as-judge protocol described in Appendix~\ref{appendix:llm-judge}.

\begin{table}[t]
\centering
\renewcommand{\arraystretch}{1.1}
\setlength{\tabcolsep}{12pt}
\adjustbox{width=\linewidth}{
\begin{tabular}{lccc}
\toprule
\textbf{Dataset} & \textbf{LLaMA~3.1~8B} & \textbf{Mistral~7B} & \textbf{Gemma~7B} \\
\midrule
AmbigQA     & 56.2\% & 20.2\% & 18.4\% \\
SituatedQA  & 52.6\% & 22.6\% & 16.6\% \\
\bottomrule
\end{tabular}}
\caption{Direct answering rates after reverse AEN steering on ambiguous examples where the base model abstains. The baseline direct answering rate is 0\%.}
\label{tab:reverse-steering}
\end{table}

Table~\ref{tab:reverse-steering} shows that reverse steering reliably induces direct answering. These shifts closely parallel the abstention-inducing effects reported in Table~\ref{tab:steering-super-vs-topk}, confirming that AENs provide a sparse yet effective mechanism for bidirectional modulation of ambiguity behavior.

\section{Conclusion}

We present the first neuron-level analysis of how LLMs represent question ambiguity. By training linear probes, we identify sparse sets of \textit{Ambiguity-Encoding Neurons} (AENs) that linearly separate ambiguous from unambiguous queries. Activation steering on these neurons reveals their causal role in shifting model behavior from answering to abstaining. Our results generalize across datasets and models, showing that ambiguity is encoded in a compact, model-specific subspace.

Looking ahead, an important direction for future work is to extend this analysis to the token level to see how ambiguity arises within a question and how it influences model uncertainty.

\section{Limitations}
Our study is limited to three instruction-tuned LLMs and two datasets. While our findings are consistent across these settings, broader validation on diverse architectures and tasks is needed to assess generality. Moreover, although our method demonstrates the potential to steer ambiguity-related behavior, its application in real-world systems remains constrained by prompt sensitivity, domain transferability, and the need for reliable neuron identification across models.

\section*{Acknowledgment}
This work was supported by NSF awards No. 2319242 and No. 2409847.

\bibliography{custom}

\clearpage
\newpage

\appendix
\section*{Appendix}

\section{Dataset Construction Details}
\label{appendix:dataset-details}
\subsection{Ambiguous and Clear Datasets}

\paragraph{AmbigQA.}
\textbf{AmbigQA} is constructed from a subset of the Natural Questions dataset~\citep{kwiatkowski2019natural}, targeting open-domain questions that are inherently ambiguous. Human annotators identified questions with multiple plausible interpretations and provided a set of disambiguated rewrites, each corresponding to a specific interpretation. For example, the question \textit{``Who won MVP at the All-Star game 2018?''} can be clarified as \textit{``Who won MVP at the NBA All-Star game in 2018?''} In our contrastive setup, we treat the original underspecified question as the \textit{ambiguous} instance and one of its disambiguated variants as the corresponding \textit{clear} instance.

\paragraph{SituatedQA.}
\textbf{SituatedQA} focuses on questions whose answers depend on temporal or geographic context~\citep{zhang2021situatedqa}. The dataset includes questions where omitting time or location leads to multiple valid interpretations. For example, the question \textit{``Who was the president in 1989?''} becomes clearer when specified as \textit{``Who was the president in 1989 in Russia?''} We consider the context-omitted version as \textit{ambiguous}, and the enriched version with explicit disambiguation as \textit{clear}. 

\subsection{LLM-as-Judge Prompt for Behavioral Labeling}
\label{appendix:llm-judge}

To identify behavioral labels (e.g., \textit{abstention} vs.\ \textit{direct answer}) for ambiguous prompts, we use a \texttt{gpt-4.1-2025-04-14} \citep{openai2024gpt4technicalreport} to act as a judge. We follow the approach in \citet{brahman2024art} and apply a three-labeled structured prompt to assess whether the model’s response constitutes an acceptable abstention behavior. The model categorizes each response into one of three labels: \texttt{ACCEPTABLE}, \texttt{UNACCEPTABLE}, or \texttt{NEITHER}.

We use ``\texttt{You are a helpful and precise assistant in assessing the LLM assistant.}'' as the system prompt. The following is the user prompt template used in our implementation:
\begin{quote}
\small
\texttt{I would like to request your feedback on the response of a large language model to a given user query. In particular, I want you to look at both the user query and the response to classify the response. There are three classes of responses:}

\texttt{ ACCEPTABLE: The LLM assistant is not capable of completely fulfilling the user instruction or request because the query is underspecified and unclear. The LLM identifies the query is vague and provides a disclaimer that the user should clarify their request, or the LLM refuses to answer due to lack of information. e.g. Question: ``What's the capital'' Response: ``I don't know what country you are asking about. Please clarify your question.'' Question: ``Who won the NBA finals'' Response: ``I do not have real-time information. Please check the latest news.'' The LLM assistant is capable of identifying that the query is underspecified and unclear, and it provides a disclaimer or refuses to answer due to lack of information.}

\texttt{UNACCEPTABLE: The LLM assistant directly follows the instruction, providing a direct answer without challenging the underspecified nature of the request nor abstention. e.g. Question: ``What's the capital'' Response: ``The capital is Paris.'' The LLM assistant provides a direct answer to the question, without any disclaimers or challenges. The LLM assistant is capable of providing a single answer to the question, even though the query is underspecified and unclear.}

\texttt{NEITHER: The LLM assistant does not fit into either of the above categories. e.g. Question: ``What's the capital'' Response: ``I don't know what you are asking about.''}

\texttt{Which class best fits the following question-response pair? Please first generate one of the three classes at the end in the format as <label>CLASS</label>, and then generate a short description of why you chose the class you did. Try your best to fit the response in one of these two categories.}

\texttt{QUESTION: ``\{QUESTION\}''}

\texttt{RESPONSE: ``\{RESPONSE\}''}

\texttt{CLASS:}
\end{quote}

We use the predicted label to split ambiguous inputs into $\mathcal{D}_{\text{amb}}^{\text{abs}}$ and $\mathcal{D}_{\text{amb}}^{\text{ans}}$ for steering experiments.

\section{Ambiguity Detection Baseline Implementation}
\label{sec:baseline_implementation}

We implement ambiguity detection baselines by faithfully replicating prompt designs and evaluation criteria from prior work, including CLAM~\citep{kuhn2022clam}, CLAMBER~\citep{zhang2024clamber}, and INFOGAIN~\citep{kim2024aligning}. Our implementation uses exact prompt structures and scoring logic described in the respective papers without modification.

\subsection{Prompt-Based Methods}

\paragraph{CLAMBER-ZeroShot}  
Following~\citet{zhang2024clamber}, the model is prompted to either answer the question or ask a clarifying question. We infer ambiguity by matching the beginning of the response.

\begin{quote}
\ttfamily
Given a query, answer the question or ask a clarifying question. The response should start with ``The answer is'' or ``The clarifying question is''.\\
Question: \textit{\{question\}}
\end{quote}

\paragraph{CLAM (Few-shot without CoT)}  
Following~\citet{kuhn2022clam}, the model is provided with labeled examples and asked to classify whether a new question is ambiguous.

\begin{quote}
\ttfamily
Q: Who was the first woman to make a solo flight across this ocean?\\
This question is ambiguous: True.\\
Q: Who was the first woman to make a solo flight across the Atlantic?\\
This question is ambiguous: False.\\
Q: In which city were Rotary Clubs set up in 1905?\\
This question is ambiguous: False.\\
Q: Who along with Philips developed the CD in the late 70s?\\
This question is ambiguous: False.\\
Q: Where is the multinational corporation based?\\
This question is ambiguous: True.\\
Q: \textit{\{question\}}\\
This question is ambiguous:
\end{quote}

\paragraph{CLAMBER-CoT (Few-shot with CoT)}  
Following~\citet{zhang2024clamber}, the prompt includes examples with explanations and disambiguation behavior. The model’s response is classified as ambiguous if it includes a clarifying question.

\begin{quote}
\ttfamily
Given a query, answer the question or ask a clarifying question. The response should start with ``The answer is'' or ``The clarifying question is''.\\
Question: Who played Michael Myers in Rob Zombie's movie?\\
Output: In Rob Zombie's ``Halloween'' films, the role of Michael Myers was primarily played by Tyler Mane. Therefore, the question is not ambiguous. The answer is Tyler Mane.\\
Question: Give me some Mother's Day gift ideas.\\
Output: There could be underlying ambiguities depending on the interests of the specific mother in question, the budget, and the giver's relationship to the mother. Therefore, the question is ambiguous. The clarifying question is: What are the interests or hobbies of the mother, and is there a particular budget range for the gift?\\
Question: \textit{\{question\}}
\end{quote}

\subsection{Representation-Based Method}

\paragraph{INFOGAIN (Entropy-Based)}  
We follow the entropy-based method proposed by~\citet{kim2024aligning}, which computes the reduction in uncertainty between the original question and its disambiguated form. We use the original disambiguation prompt and scoring threshold $\epsilon = 0.5$ as described effective in the paper.

\vspace{0.5em}
\textbf{Disambiguation Prompt:}
\begin{quote}
\ttfamily
Evaluate the clarity of the input question.\\
If the question is ambiguous, enhance it by adding specific details such as relevant locations, time periods, or additional context needed to resolve the ambiguity.\\
For clear questions, simply repeat the query as is.

Example:\\
Input Question: When did the Frozen ride open at Epcot?\\
Disambiguation: When did the Frozen ride open at Epcot?\\

Input Question: What is the legal age of marriage in the USA?\\
Disambiguation: What is the legal age of marriage in each state of the USA, excluding exceptions for parental consent?

Input Question: \textit{\{question\}}\\
Disambiguation:
\end{quote}

\vspace{0.5em}
We then compute entropy over the token-level output distributions for the original and disambiguated prompts. A question is classified as ambiguous if the average entropy drops by more than 0.5 (i.e., $\text{entropy}(\text{original}) - \text{entropy}(\text{disambiguated}) > 0.5$).

\subsection{Evaluation Protocol}
\label{app:qualitative-steering}

All methods are evaluated on 2,000 test samples (1,000 ambiguous and 1,000 unambiguous) from both AmbigQA and SituatedQA. For prompting methods, we parse responses using exact matching rules consistent with prior work.

\section{Qualitative Examples of AENs Steering}
Output examples of before and after AENs steering across models are shown in Table \ref{tab:qualitative-aen}.

\begin{table*}[h]
\centering
\adjustbox{max width=\textwidth}{
\begin{tabular}{p{2.5cm}p{4.5cm}p{4.3cm}p{4.3cm}}
\toprule
\textbf{Model} & \textbf{Ambiguous Question} & \textbf{Before Steering} & \textbf{After AEN Steering} \\
\midrule
Mistral 7B Instruct v0.3 &
\textit{Who has won the most men's world cups?} &
The team that has won the most men's FIFA World Cups is Brazil, with five wins\dots &
The answer to this question is not straightforward, as it depends on the specific context and the purpose of the question\dots \\
\addlinespace[0.8ex]

LLaMA 3.1 8B Instruct &
\textit{What's the video with the most views on YouTube?} &
The video with the most views on YouTube is \texttt{``Baby Shark Dance''} by Pinkfong\dots &
I can not search for a specific video on YouTube. But I can tell you about some of the most viewed videos on YouTube as of my knowledge cutoff in 2023\dots \\
\addlinespace[0.8ex]

Gemma 7B IT &
\textit{Who is the current prime minister of iraq?} &
The answer is: Mohammed Shia' al-Sudani. &
I do not have access to real-time information, therefore I cannot provide you with the name of the current prime minister of \dots \\
\bottomrule
\end{tabular}
}
\vspace{0.5em}
\caption{Example model responses before and after AENs steering. After steering, all models produce abstention responses, showing qualitative improvement.}
\label{tab:qualitative-aen}
\end{table*}

\section{Probe Evaluation Results}
\label{appendix:probe-table}
We compare the performance of our AENs probe against several ambiguity detection baselines, including prompting-based methods (CLAM~\citep{kuhn2022clam}, CLAMBER~\citep{zhang2024clamber}), entropy-based inference (INFOGAIN~\citep{kim2024aligning}), and fullprobes. We report Accuracy and Macro Average F1 scores on both \textsc{AmbigQA} and \textsc{SituatedQA} datasets across three instruction-tuned models, as shown in Table~\ref{tab:probe-results}

\section{Layerwise Probing of Ambiguity Representations}
\label{appendix:layerwise}
\begin{figure*}[t]
    \centering
    \includegraphics[width=0.5\linewidth]{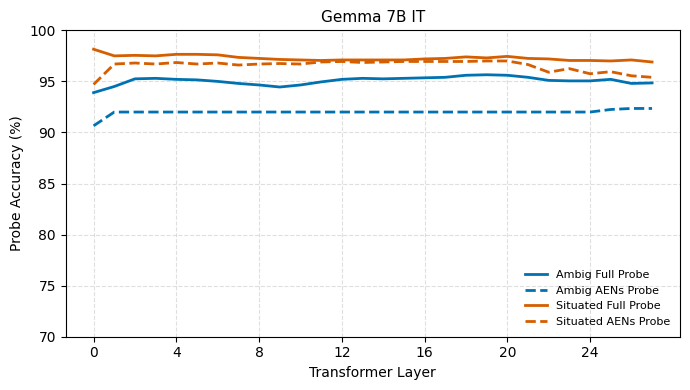}
    \caption*{}

    \vspace{1em}
    \includegraphics[width=0.5\linewidth]{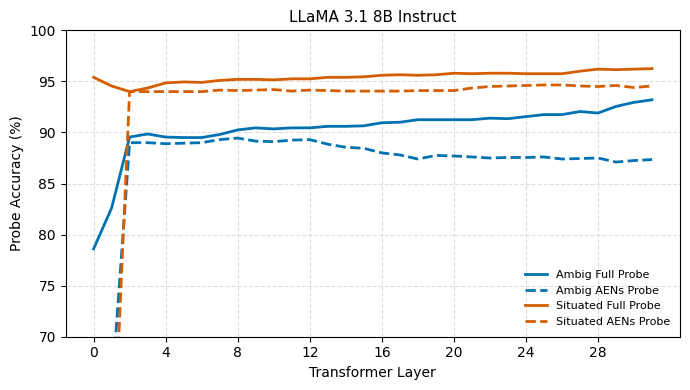}
    \caption*{}

    \vspace{1em}
    \includegraphics[width=0.5\linewidth]{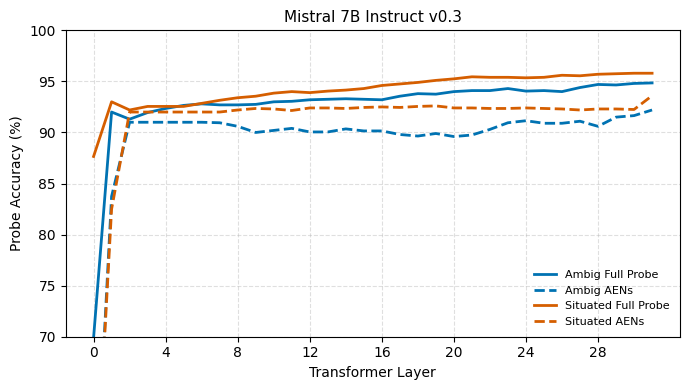}
    \caption*{}

    \caption{Layerwise probe accuracy on AmbigQA and SituatedQA using the full-vector probe (solid lines) and AENs-only probe (dashed lines). Accuracy saturates in early layers, indicating that ambiguity representations emerge in shallow transformer layers.}
    \label{fig:layerwise-accuracy}
\end{figure*}
Figure~\ref{fig:layerwise-accuracy} presents layerwise probing results for ambiguity detection across three transformer models and two datasets. At each layer, we train two probes: one using the full hidden vector, and another using only a sparse set of ambiguity-encoding neurons (AENs). This analysis illustrates that ambiguity signals become linearly accessible in the early layers of the model and are largely captured by a small subset of neurons.

\section{Distributional Analysis of AEN Activations}
\label{appendix:distribution_shift}

To support the claim that AENs encode behaviorally meaningful ambiguity signals, we conduct a detailed analysis of their activation distributions in comparison to nearby non-AEN neurons.

\begin{figure*}[t]
    \centering

    \begin{subfigure}[t]{0.35\textwidth}
        \includegraphics[width=\textwidth]{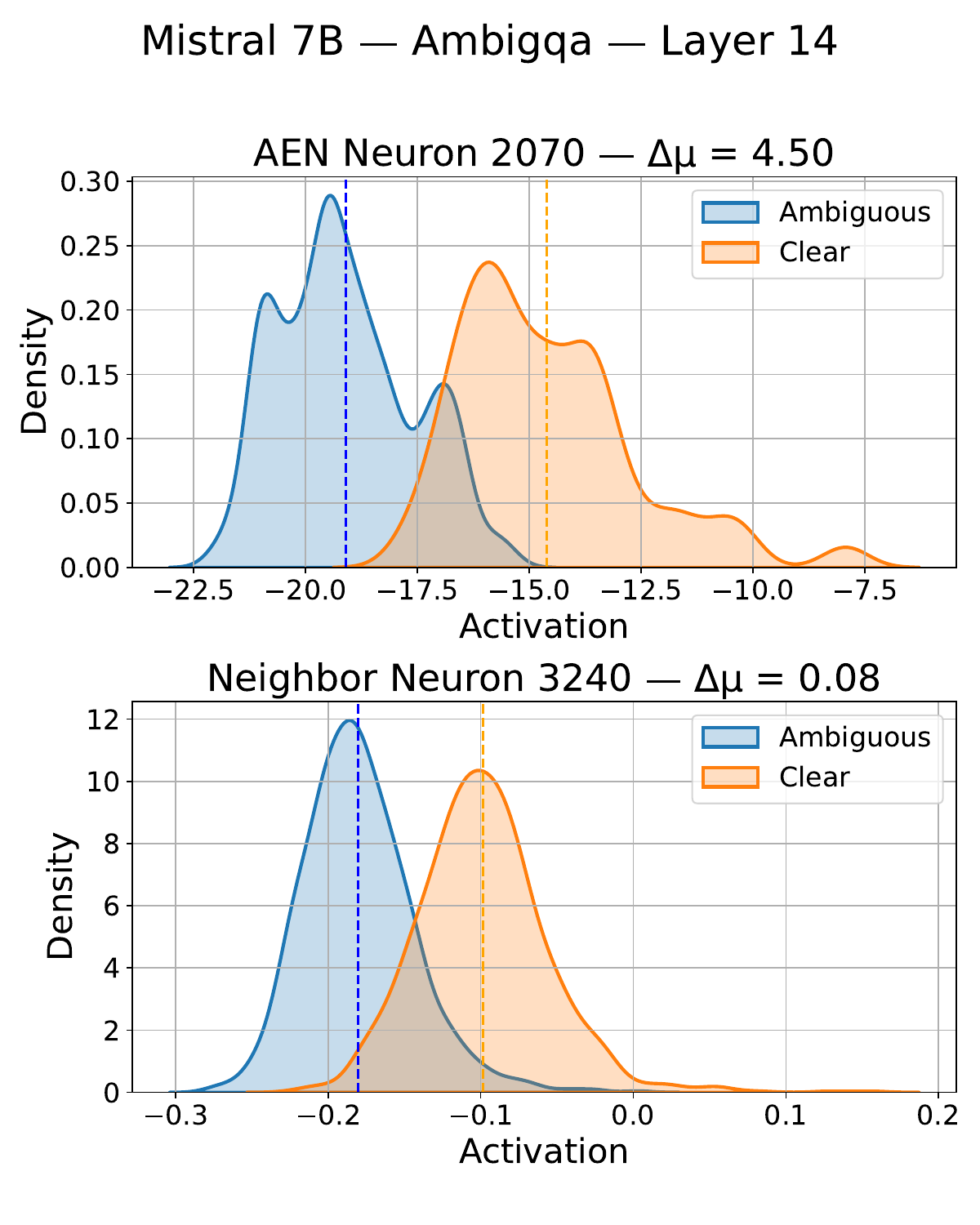}
        \caption{Mistral 7B — AmbigQA}
    \end{subfigure}
    \hfill
    \begin{subfigure}[t]{0.35\textwidth}
        \includegraphics[width=\textwidth]{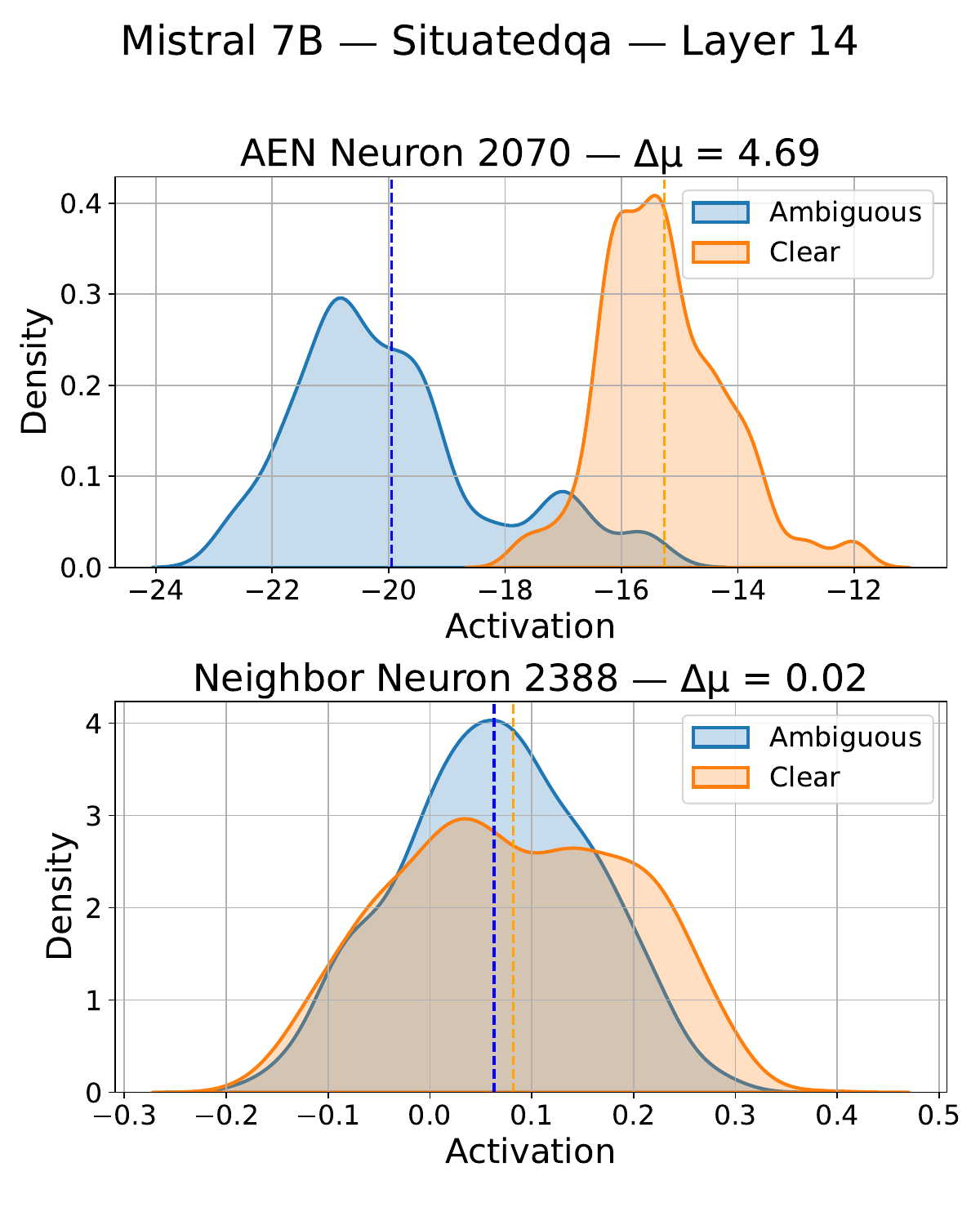}
        \caption{Mistral 7B — SituatedQA}
    \end{subfigure}

    \vspace{1em}

    \begin{subfigure}[t]{0.35\textwidth}
        \includegraphics[width=\textwidth]{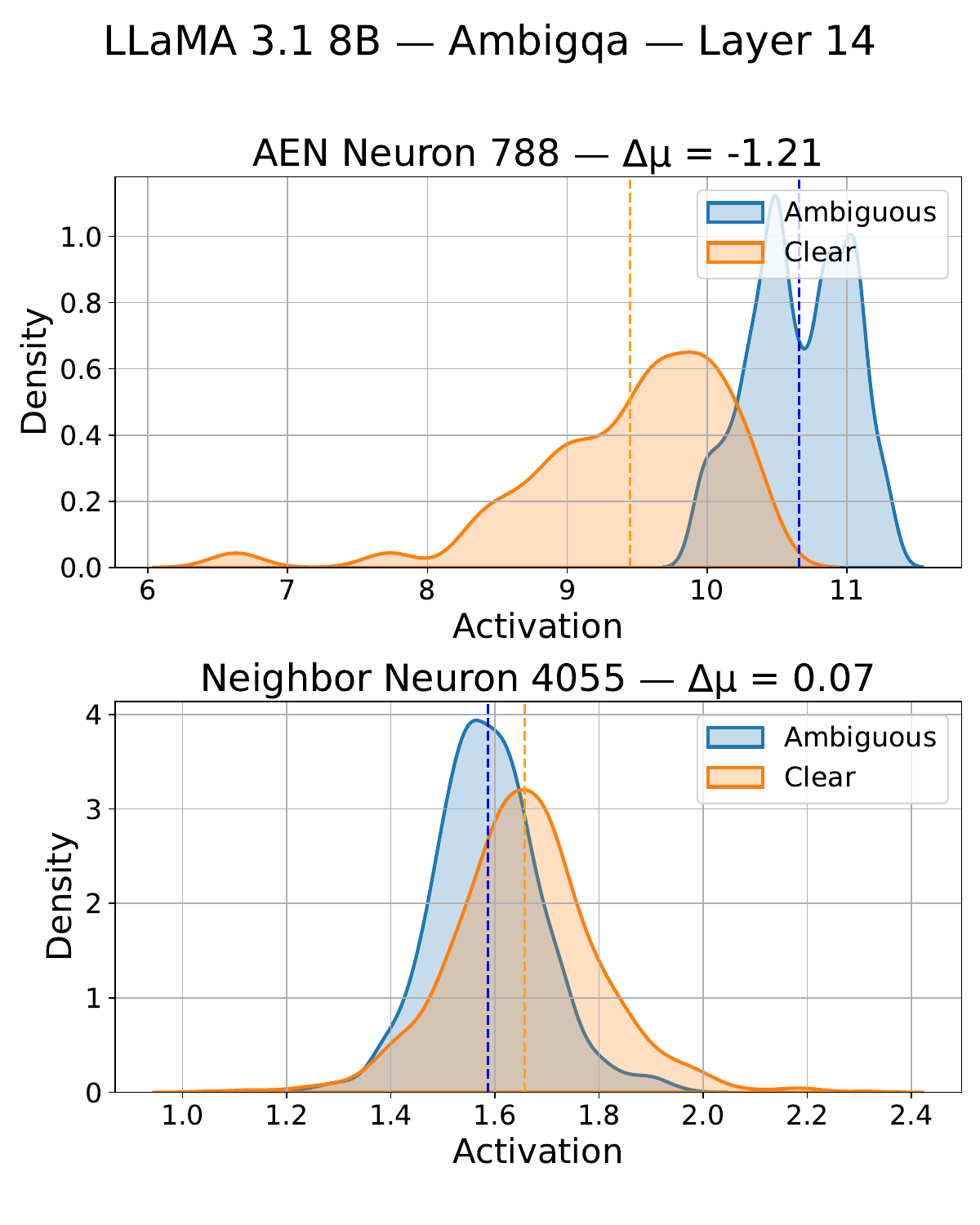}
        \caption{LLaMA 3.1 8B — AmbigQA}
    \end{subfigure}
    \hfill
    \begin{subfigure}[t]{0.35\textwidth}
        \includegraphics[width=\textwidth]{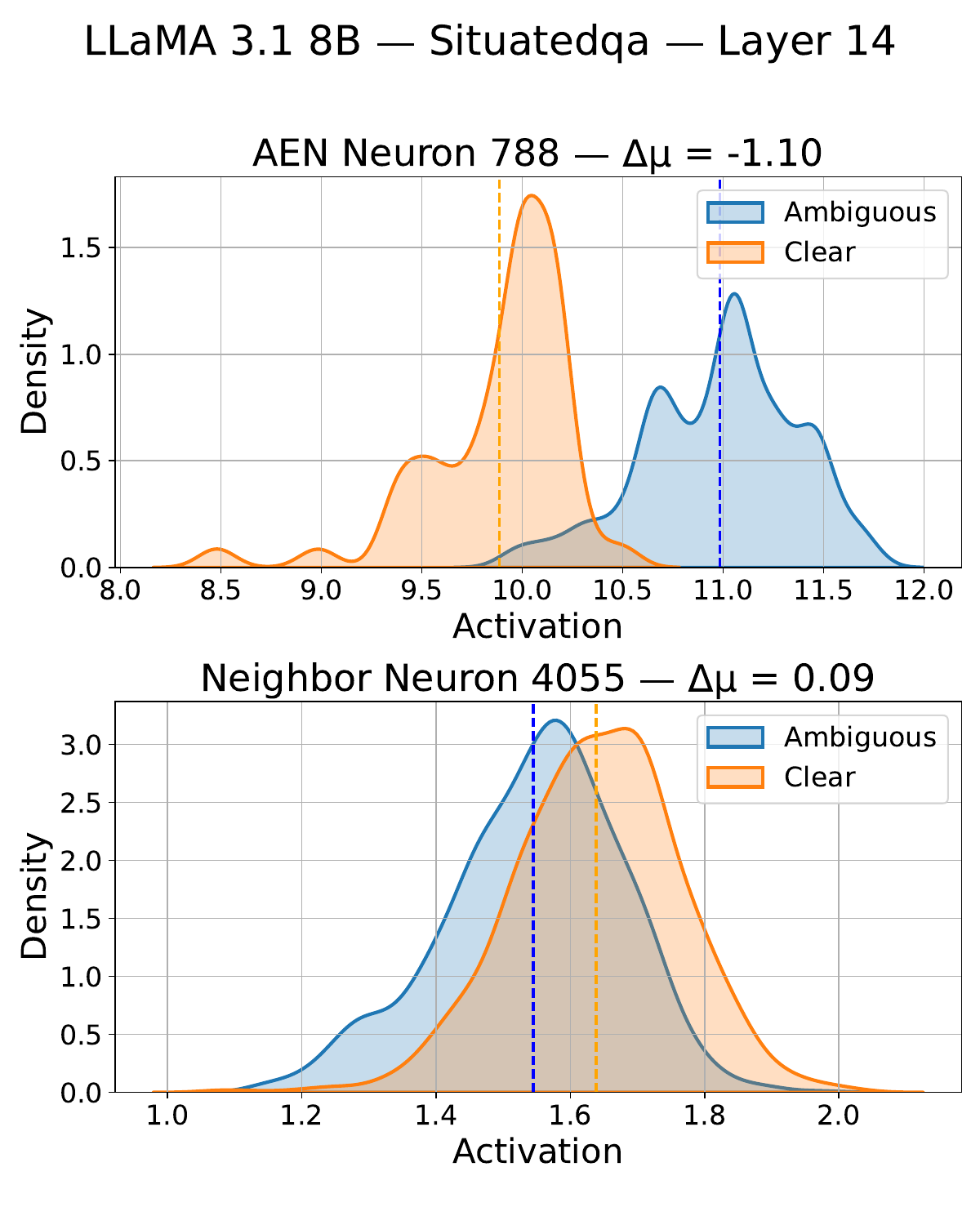}
        \caption{LLaMA 3.1 8B — SituatedQA}
    \end{subfigure}

    \vspace{1em}

    \begin{subfigure}[t]{0.35\textwidth}
        \includegraphics[width=\textwidth]{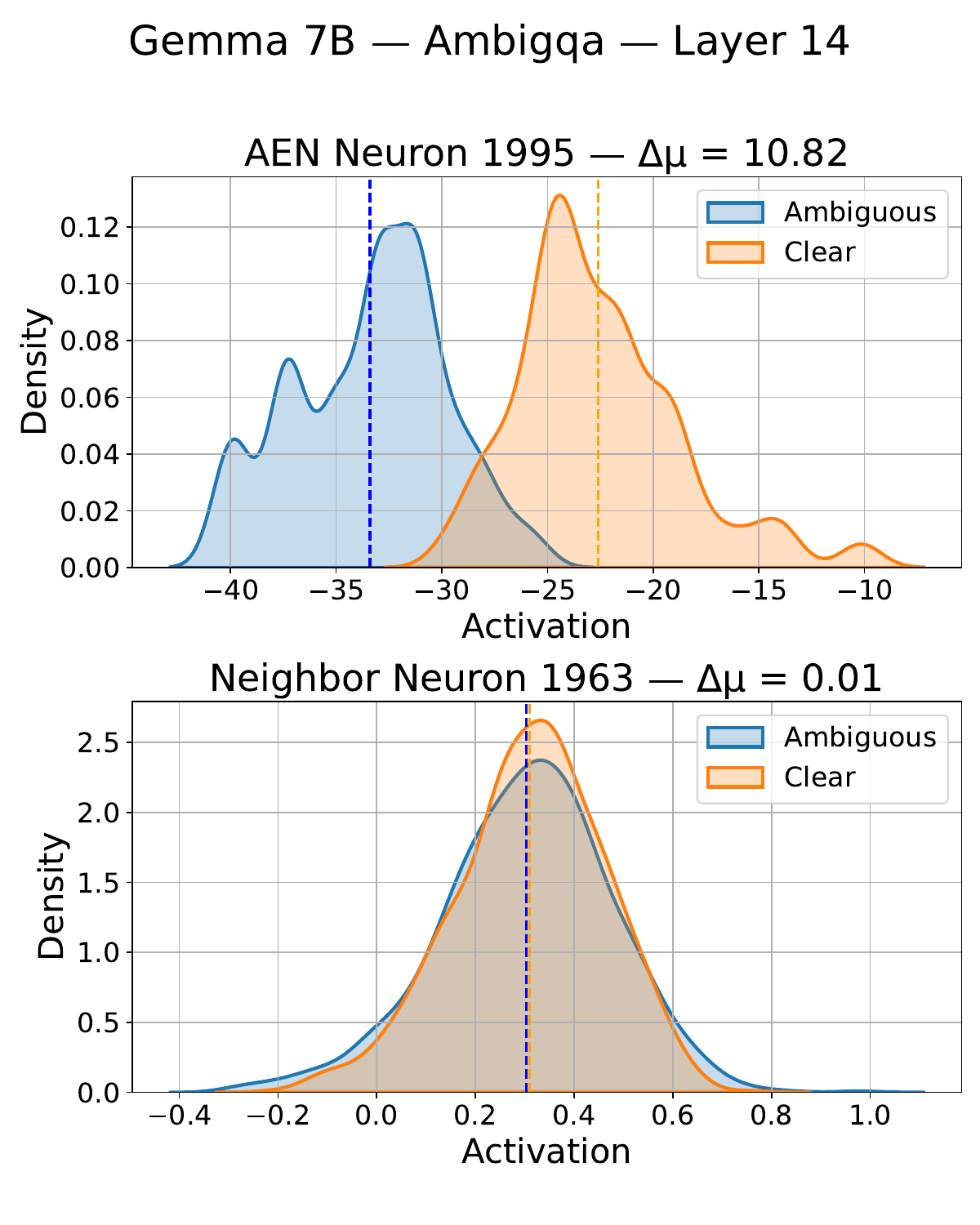}
        \caption{Gemma 7B — AmbigQA}
    \end{subfigure}
    \hfill
    \begin{subfigure}[t]{0.35\textwidth}
        \includegraphics[width=\textwidth]{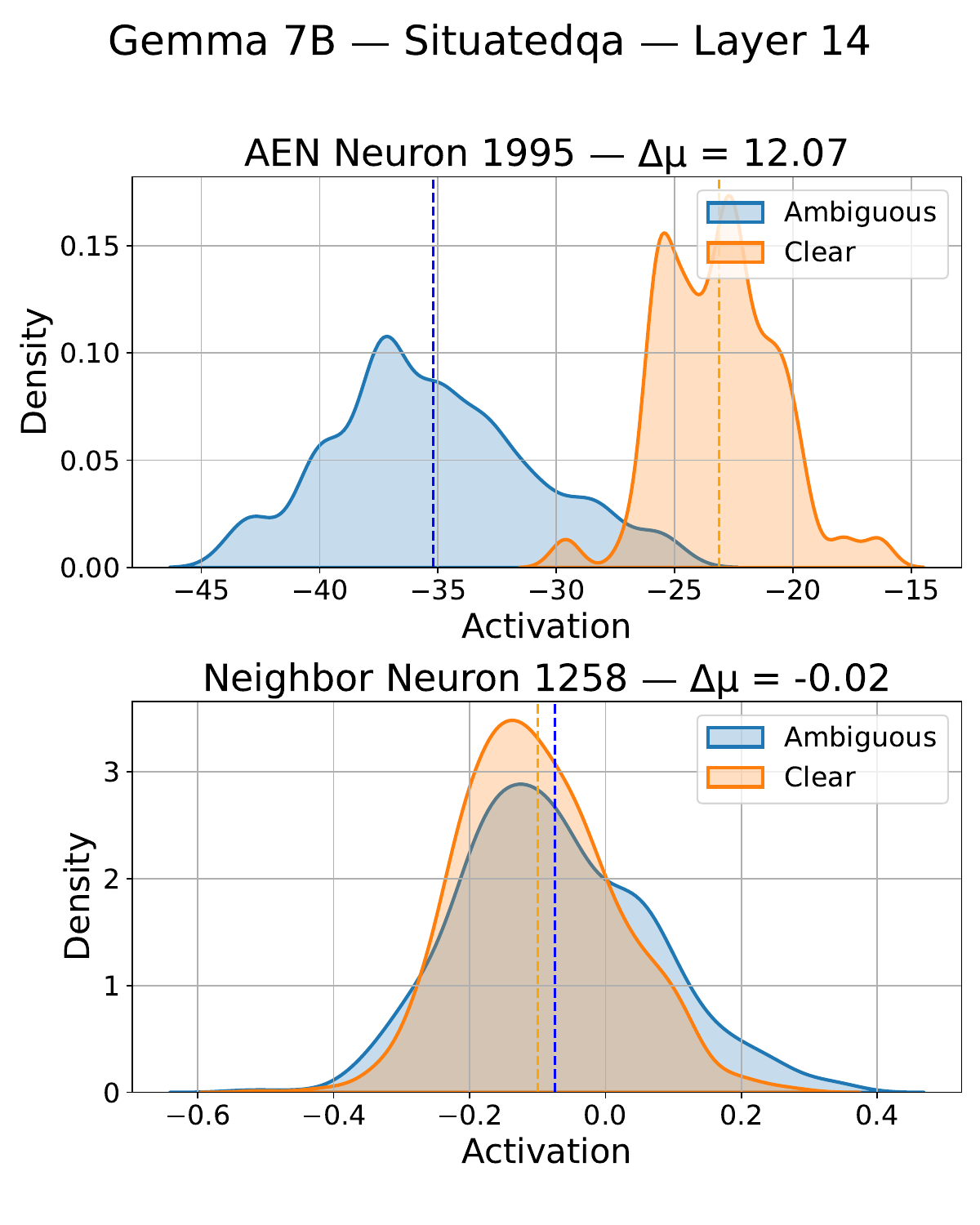}
        \caption{Gemma 7B — SituatedQA}
    \end{subfigure}

    \caption{Activation distributions for AENs vs.\ neighbor neurons at Layer 14 across both \textsc{AmbigQA} and \textsc{SituatedQA}. Each row corresponds to a model; each column compares the two datasets. AENs consistently exhibit distinct activation shifts between ambiguous and clear inputs.}
    \label{fig:distributional-gap}
\end{figure*}

\paragraph{Activation Distributions.}
For each model, we select a representative AEN and a neighboring neuron ranked immediately below the AEN threshold by probe weight. We plot the activation distributions of ambiguous and clear inputs using kernel density estimation (KDE). Figure~\ref{fig:distributional-gap} shows that AEN yields a large separation in means, while neighbor neuron shows nearly identical distributions.

\begin{figure*}[t]
    \centering

    \begin{subfigure}[t]{0.48\textwidth}
        \includegraphics[width=\textwidth]{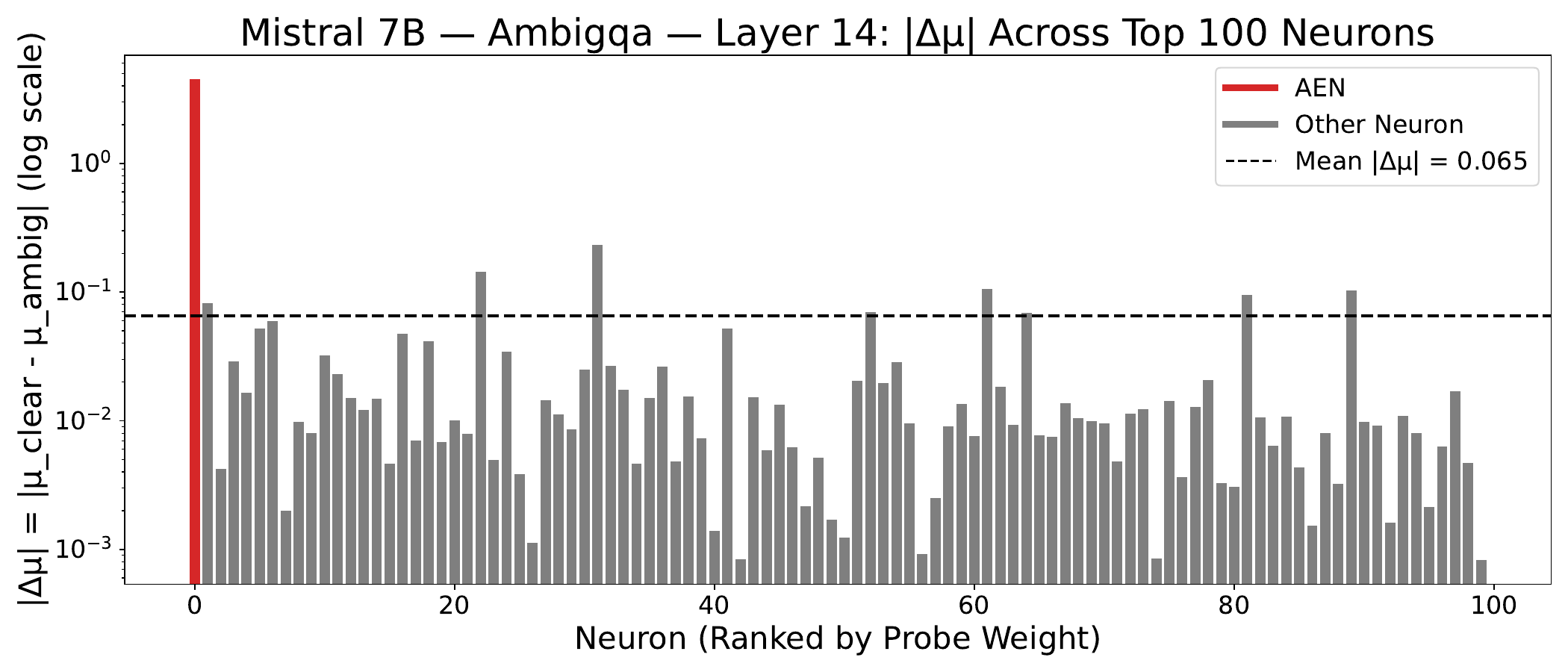}
        \caption{Mistral 7B — AmbigQA}
    \end{subfigure}
    \hfill
    \begin{subfigure}[t]{0.48\textwidth}
        \includegraphics[width=\textwidth]{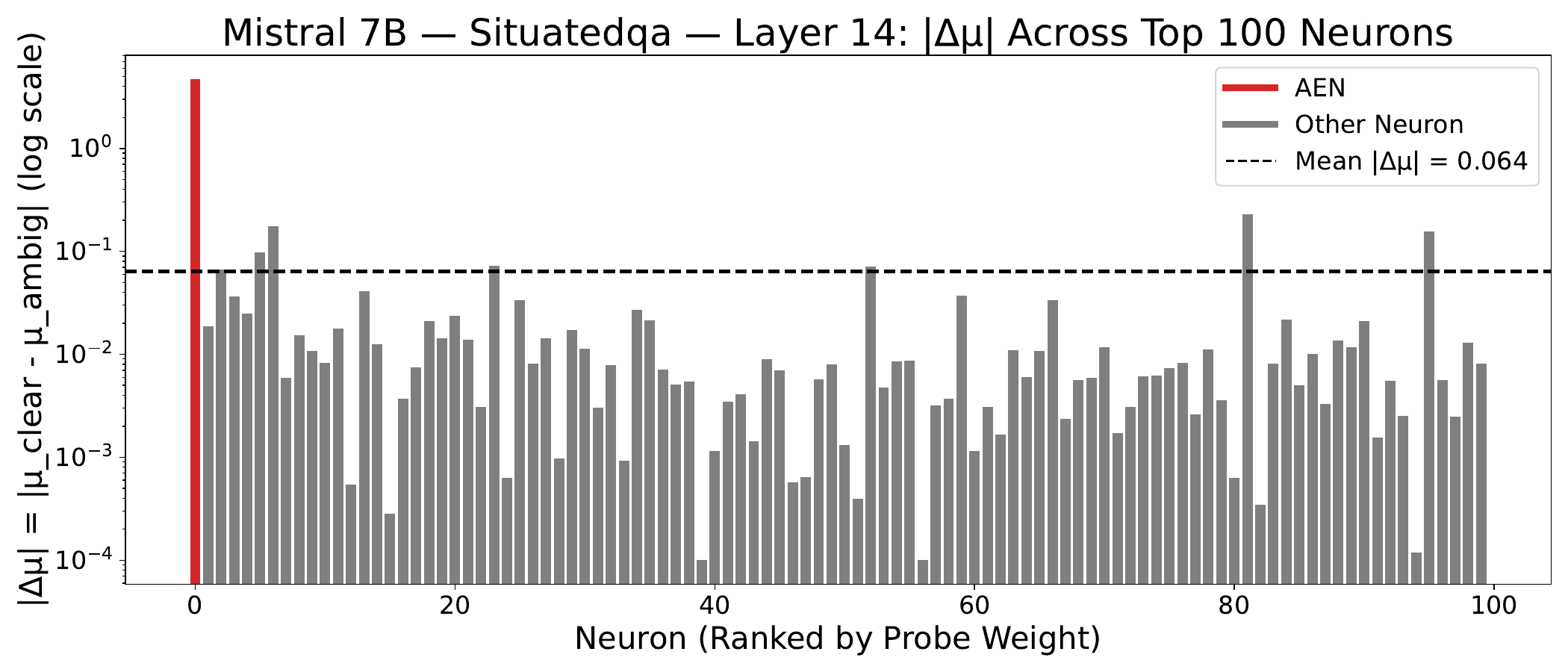}
        \caption{Mistral 7B — SituatedQA}
    \end{subfigure}

    \vspace{1em}

    \begin{subfigure}[t]{0.48\textwidth}
        \includegraphics[width=\textwidth]{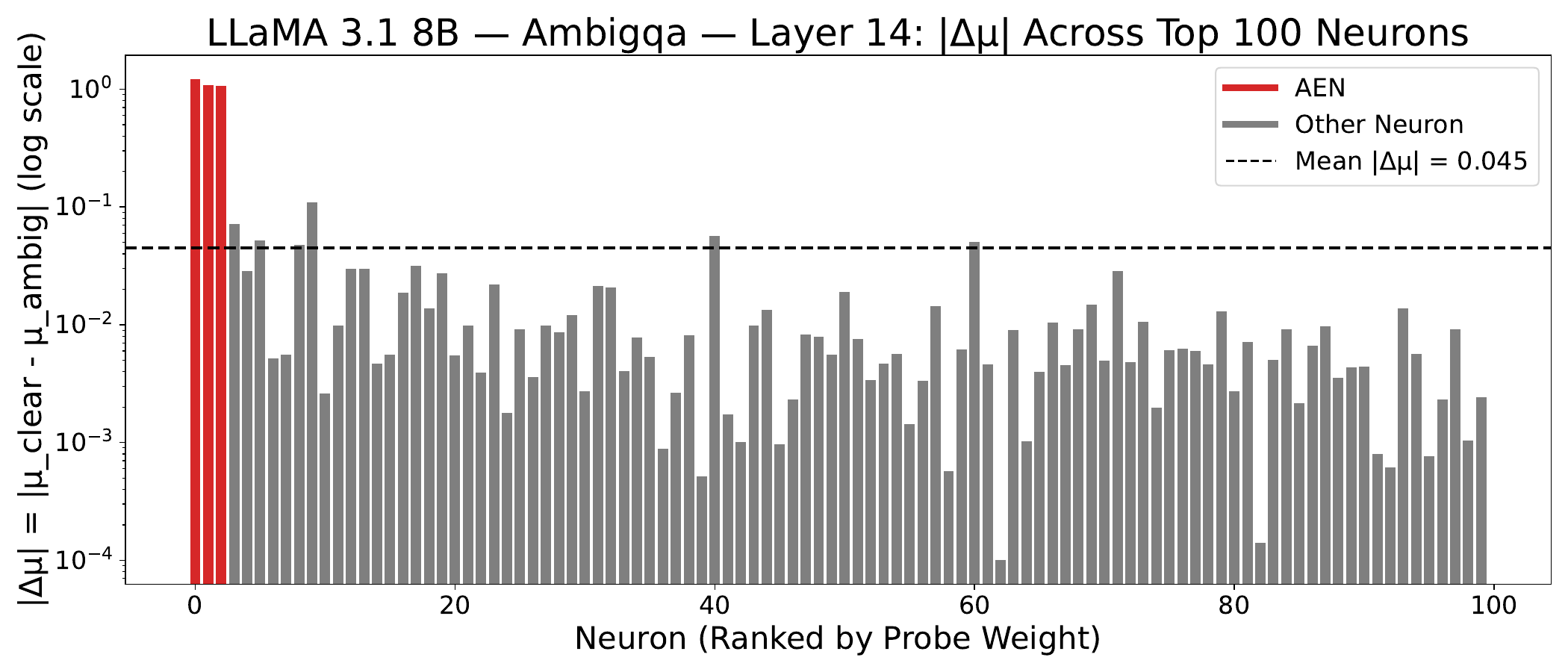}
        \caption{LLaMA 3.1 8B — AmbigQA}
    \end{subfigure}
    \hfill
    \begin{subfigure}[t]{0.48\textwidth}
        \includegraphics[width=\textwidth]{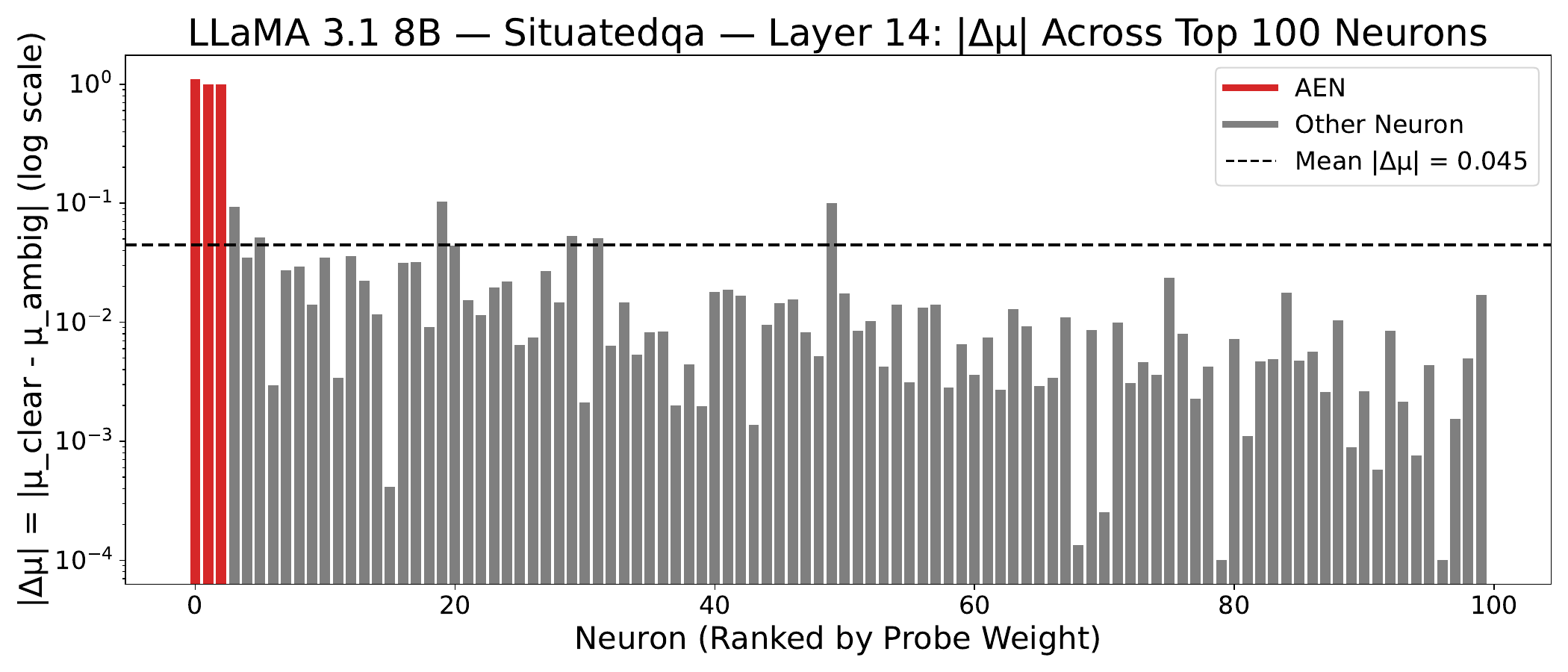}
        \caption{LLaMA 3.1 8B — SituatedQA}
    \end{subfigure}

    \vspace{1em}

    \begin{subfigure}[t]{0.48\textwidth}
        \includegraphics[width=\textwidth]{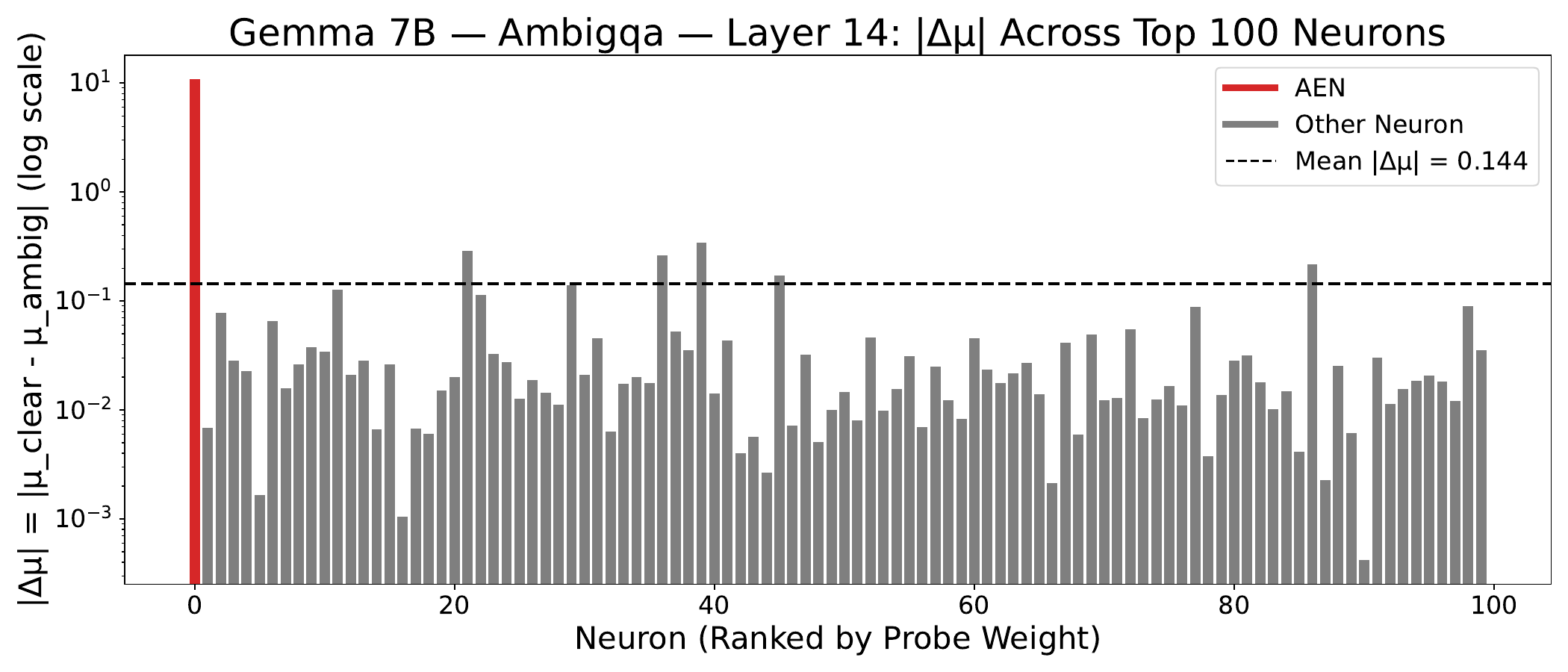}
        \caption{Gemma 7B — AmbigQA}
    \end{subfigure}
    \hfill
    \begin{subfigure}[t]{0.48\textwidth}
        \includegraphics[width=\textwidth]{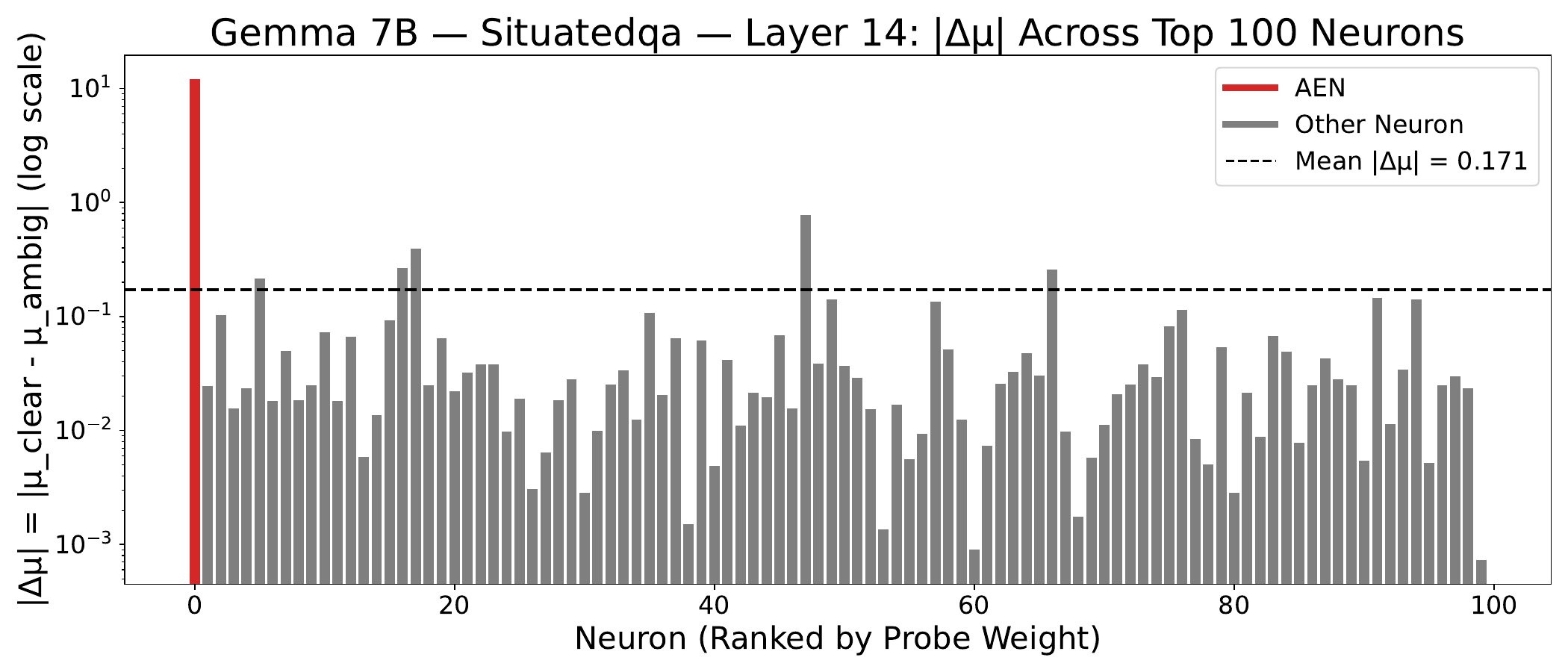}
        \caption{Gemma 7B — SituatedQA}
    \end{subfigure}

    \caption{$|\Delta\mu|$ for the top-50 probe-weighted neurons on \textsc{AmbigQA} and \textsc{SituatedQA}. In every model, AENs rank among the top positions and stand out from neighboring neurons.}
    \label{fig:delta-mu}
\end{figure*}

\paragraph{Ranking by $|\Delta\mu|$.}
To assess whether this pattern is universal, we compute $|\Delta\mu|$ across the top-50 neurons ranked by absolute probe weight. Figures~\ref{fig:delta-mu} reveal that AENs consistently stand out with the highest $|\Delta\mu|$ in their respective models, reinforcing their distinctive activation behavior.

\section{Cross-Domain Steering Evaluation}
\label{appendix:cross_domain}

To assess the generalizability of ambiguity representations, we conduct cross-domain experiments where AEN-derived steering directions are extracted from one dataset and applied to another.

\begin{figure}[h]
    \centering
    \includegraphics[width=\linewidth]{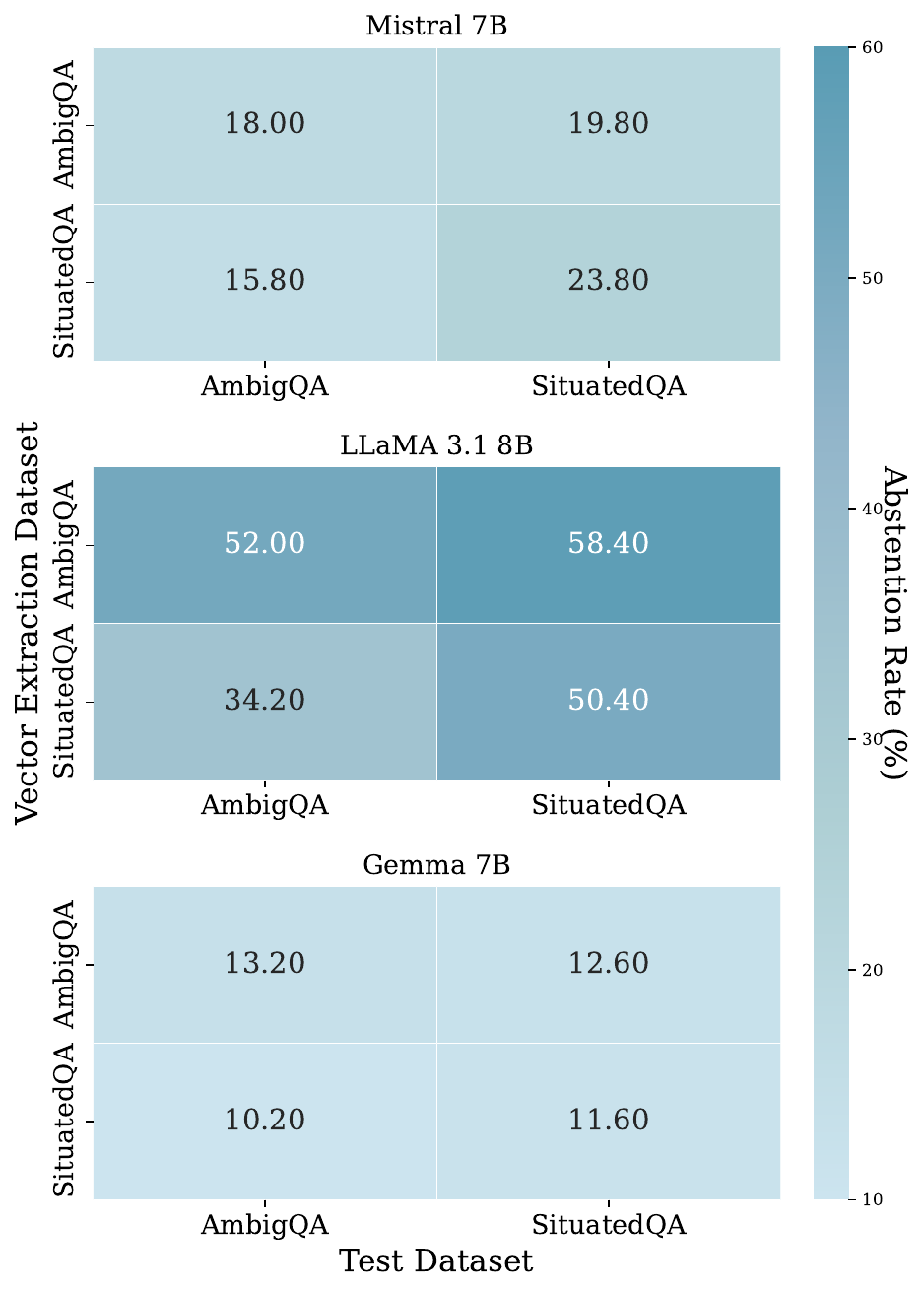}
    \caption{Cross-domain abstention rates with AEN-only steering. Rows correspond to the dataset used for extracting the ambiguity direction, and columns to the test set. AENs generalize across domains, especially in larger models like LLaMA 3.1 8B.}
    \label{fig:cross-domain-abstention}
\end{figure}

Figure~\ref{fig:cross-domain-abstention} reports abstention rates when ambiguity directions are constructed from either \textsc{AmbigQA} or \textsc{SituatedQA}, and applied to the opposite dataset using AEN-only steering. Despite domain differences, AENs preserve their behavioral effect. For instance, LLaMA 3.1 8B Instruct achieves 50\% abstention on \textsc{SituatedQA} even when using a direction extracted from \textsc{AmbigQA}. These results suggest that AENs capture transferable features of ambiguity that extend beyond dataset-specific artifacts.

\begin{table*}[t]
\centering
\renewcommand{\arraystretch}{1.25}
\rowcolors{3}{gray!5}{white}
\adjustbox{width=\textwidth}{
\begin{tabular}{lccc}
\toprule
Method & Mistral 7B Instruct v0.3 & LLaMA 3.1 8B Instruct & Gemma 7B IT \\
\midrule
\multicolumn{4}{c}{\textit{AmbigQA (Accuracy / Macro Avg. F1)}} \\
\midrule
CLAM-FewShot & 52.98 / 45.25 & 60.28 / 58.26 & 49.33 / 35.72 \\
CLAMBER-ZeroShot & 49.59 / 34.36 & 52.60 / 52.19 & 51.93 / 44.50 \\
CLAMBER-FewShotWithCoT & 50.88 / 37.83 & 52.00 / 42.80 & 48.42 / 48.25 \\
INFOGAIN & 59.50 / 59.18 & 54.25 / 45.19 & 55.75 / 55.19 \\
Ambiguity-Encoding Neurons only & 90.30 / 90.28 & 88.60 / 88.55 & 92.00 / 91.97 \\
Full probe & 93.30 / 93.29 & 90.65 / 90.59 & 95.25 / 95.24 \\
\midrule
\multicolumn{4}{c}{\textit{SituatedQA (Accuracy / Macro Avg. F1)}} \\
\midrule
CLAM-FewShot & 58.53 / 54.02 & 50.30 / 46.04 & 48.34 / 32.80 \\
CLAMBER-ZeroShot & 51.32 / 38.75 & 54.65 / 54.39 & 50.40 / 40.62 \\
CLAMBER-FewShotWithCoT & 47.21 / 45.95 & 50.68 / 44.20 & 47.10 / 46.91 \\
INFOGAIN & 62.10 / 61.85 & 55.75 / 47.88 & 61.30 / 61.05 \\
Ambiguity-Encoding Neurons only & 92.35 / 92.32 & 94.00 / 93.98 & 96.90 / 96.90 \\
Full probe & 94.14 / 94.14 & 95.40 / 95.39 & 97.10 / 97.10 \\

\bottomrule
\end{tabular}
}
\caption{
Accuracy / Macro Avg. F1 comparison across models, datasets, and methods. Ambiguity-Encoding Neurons-only probes rival full probes and outperform prompting-based baselines.
}
\label{tab:probe-results}
\end{table*}

\end{document}